\documentclass[3p,authoryear,12pt]{elsarticle}
\usepackage{subfigure}
\usepackage{epsfig}
\usepackage{amsmath}
\usepackage{amssymb}
\usepackage[boxed, lined]{algorithm2e}
\usepackage{subfigure}
\usepackage{tikz}
\usepackage{xspace}
\usepackage{url}

\usepackage{makeidx}
\usepackage{graphicx}
\usepackage{lineno}
\usepackage{booktabs}
\usepackage{hhline}
\usepackage{gensymb}

\newcommand{\ie}{i.e.~}

\newcommand{\eg}{e.g.~}

\renewcommand{\(}{\left(}
\renewcommand{\)}{\right)}
\renewcommand{\d}[1]{{\mbox{\boldmath$#1$}}}

\newcommand{\m}[1]{{\mbox{{\fontencoding{T1}\sffamily\slshape{#1\/}}}}}

\newcommand{\trans}[0]{^{\sf T}}

\newcommand{\vlabels}{\d y}

\newcommand*\patchAmsMathEnvironmentForLineno[1]{%
\expandafter\let\csname old#1\expandafter\endcsname\csname #1\endcsname
\expandafter\let\csname oldend#1\expandafter\endcsname\csname end#1\endcsname
\renewenvironment{#1}%
         {\linenomath\csname old#1\endcsname}%
         {\csname oldend#1\endcsname\endlinenomath}}%
\newcommand*\patchBothAmsMathEnvironmentsForLineno[1]{%
      \patchAmsMathEnvironmentForLineno{#1}%
      \patchAmsMathEnvironmentForLineno{#1*}}%
    \AtBeginDocument{%
    \patchBothAmsMathEnvironmentsForLineno{equation}%
    \patchBothAmsMathEnvironmentsForLineno{align}%
    \patchBothAmsMathEnvironmentsForLineno{flalign}%
    \patchBothAmsMathEnvironmentsForLineno{alignat}%
    \patchBothAmsMathEnvironmentsForLineno{gather}%
    \patchBothAmsMathEnvironmentsForLineno{multline}%
}

\graphicspath{{images_for_submission_pdf/}}
 
\usepackage{fancyhdr}    
\fancypagestyle{firstpage}{%
  \lhead{Computers and Electronics in Agriculture 100 (2014), 148--158}
}

\usepackage{lipsum}
\makeatletter
\def\ps@pprintTitle{%
 \def\@oddhead{}%
 \let\@evenhead\@empty
 \def\@oddfoot{http://dx.doi.org/10.1016/j.compag.2013.11.008}%
 \let\@evenfoot\@oddfoot}
\makeatother
    
\begin{document}
\journal{Computers and Electronics in Agriculture}

\begin{frontmatter}

\title{Automated Image Analysis Framework for the High-Throughput Determination of Grapevine Berry Sizes Using Conditional Random Fields}

\thispagestyle{firstpage}

\author[label1]{Ribana~Roscher}
\ead{rroscher@uni-bonn.de}
\author[label2]{Katja~Herzog\corref{cor1}}
\ead{katja.herzog@jki.bund.de}
\author[label1]{Annemarie~Kunkel}
\ead{s6ankunk@uni-bonn.de}
\author[label2]{Anna~Kicherer}
\ead{anna.kicherer@jki.bund.de}
\author[label2]{Reinhard~T\"opfer}
\ead{reinhard.toepfer@jki.bund.de}
\author[label1]{Wolfgang~F\"orstner}
\ead{wf@ipb.uni-bonn.de}

\cortext[cor1]{Corresponding Author. Tel.: +49 6345-41-119; Fax: +49 6345 919050 }

\address[label1]{Department of Photogrammetry, Institute for Geodesy and Geoinformation, University of Bonn, Nu\ss allee 15, 53115 Bonn, Germany}
\address[label2]{Julius K\"uhn-Institute -- Federal Research Centre for Cultivated Plants, Institute for Grapevine Breeding Geilweilerhof, 76833 Siebeldingen, Germany}

\begin{abstract}
The berry size is one of the most important fruit traits in grapevine breeding. 
Non-invasive, image-based phenotyping promises a fast and precise method for the monitoring of the grapevine berry size. 
In the present study an automated image analyzing framework was developed in order to estimate the size of grapevine berries from images in a high-throughput manner. 
The framework includes (i) the detection of circular structures which are potentially berries and (ii) the classification of these into the class 'berry' or 'non-berry' by utilizing a conditional random field. 
The approach used the concept of a one-class classification, since only the target class 'berry' is of interest and needs to be modeled. 
Moreover, the classification was carried out by using an automated active learning approach, \ie no user interaction is required during the classification process and in addition, the process adapts automatically to changing image conditions, \eg illumination or berry color. 
The framework was tested on three datasets consisting in total of 139 images. 
The images were taken in an experimental vineyard at different stages of grapevine growth according to the BBCH scale. 
The mean berry size of a plant estimated by the framework correlates with the manually measured berry size by $0.88$.

\newpage
\end{abstract}
%

\begin{keyword}
grapevine, phenotyping, berry size, images, conditional random fields, machine vision
\end{keyword}

\end{frontmatter}

\section{Introduction}
Grapevine (V.vinifera L. subsp. vinifera) is one of the oldest and one of the economically most important fruit crops. 
Grapevines are highly susceptible to various diseases like powdery and downy mildew requiring high plant protection efforts. 
Hence, grapevine breeders around the world select for high disease resistance, climatically well adapted and high quality new cultivars (\cite{Topfer2011}).
Due to the specific cultivation of grapevines as a perennial plant \eg fruit traits can only be evaluated in the vineyard and are highly influenced by environmental factors. 
Their evaluation requires several repetitions. 
Up to now phenotyping of grapevines in vineyards has been carried out by estimation applying the BBCH scale (\cite{Bloesch2008}) or OIV descriptors (\cite{OIV2001}).
It is very time consuming, requires a lot of expertise and is expensive. 
The resulting data are subjective which make subsequent analyses more difficult like the identification of new Quantitative Trait Loci (QTL). 
Accurate phenotyping is the key tool for future plant breeding. 
Objectivity, automation and precision of phenotypic data evaluation are crucial in order to reduce the phenotyping bottleneck. 

The application of digital image analysis tools and image interpretation techniques promise a technology for high-throughput phenotyping in order to (a) increase the quantity of phenotyping samples, (b) to improve the quality of recording and (c) minimize error variation.
Low-level analysis tasks such as finding geometric objects (\eg \cite{Peng2007, Chan2005}) as well as tasks with introduced semantic higher-level information have been dealt within the literature for various applications.
Especially, higher-level knowledge about the context and the spatial arrangement of objects have been early proved beneficial for object detection or semantic image segmentation (\eg \cite{Bar1996, Biederman1982, Palmer1975}).
A well established way to incorporate this knowledge is the utilization of a conditional random field, which was introduced by \cite{Lafferty2001}. 
It has been used for example by \cite{Gould2008}, \cite{Galleguillos2008} as well as \cite{Rabinovich2007} in order to incorporate semantic context between detected objects of different pre-defined classes.
Another approach was applied by \cite{Lafarge2010} or \cite{Descombes2009}, who extract different kinds of geometric objects with point processes yielding an optimal object configuration.
Such approaches assume that the objects are disconnected from each other and the background is distinct enough so that the objects are clearly visible (\cite{Lempitsky2010}).
This situation is not always given, even less for phenotyping in the field.

One challenge in digital image analysis for high-throughput phenotyping is that only one target class, such as 'berry', is of interest. 
Other classes, which are necessary for multi-class classification, are hard to gather and cannot be specified in many cases due to their high intra- and inter-class variety. 
In order to overcome this problem, the concept of one-class classification has been introduced, which distinguishes one target class from all other classes without explicitly defining them (\eg \cite{Khan2010, Tax2001, Moya1993}).  
In this framework, both conditional random fields and an one-class classifier are combined in order to find objects which belong to the target class 'berry'.
Similar to \cite{Song2013}, who are using a conditional random fields in order to model temporal dependencies in an one-class dataset, this framework exploits information of the spatial arrangement of berries in clusters.
Moreover, the framework uses an active learning approach (\cite{Settles2010}) which defines the one-class dataset from scratch in each image. 
This has the advantage that no human user interaction is required during classification process and in addition, the process adapts automatically to changing conditions, \eg illumination or berry color. 

Image-based detection of grapes is known from precision viticulture. 
For example, \cite{Nuske2011} detect and count berries for yield estimation, \cite{Berenstein2010} detect and localize berry clusters for selective spraying or \cite{Mazzetto2010} monitor canopy health and vigour utilizing optical and analogue sensors.
Image-based phenotyping in vineyards in order to support the identification of new molecular marker for grapevine breeding comprises more detailed detection and survey of small structures, \eg single grapevine berries. 
The grapevine berry size is one of the most important target fruit traits in viticulture (\cite{Fanizza2005,Cabezas2006,Costantini2008}), whereas grapevine cultivars should preferentially have uniformity size of berries (\cite{Beslic2009}). 
In general, the berry diameter is estimated by experts applying the OIV descriptor number 221 (\cite{OIV2001}). 
This descriptor enables the classification of the berry size into five classes (class 1: very narrow berries up to about $8$~mm; class 2: narrow berries about $13$~mm; class 3: medium berries about $18$~mm; class 4: wide berries about $23$~mm; and class 5: very wide berries about $28$~mm and more). 
The results of the visual estimated berry diameter by humans are subjective resulting in error variations between the results of different people. 
In addition, precision from only $5$~mm could be achieved, which is too inaccurate for precise berry size QTL calculations.
Moreover, it should be noted that the manual estimation of sufficient amounts is very time consuming and consequently the classification of the berry size is only feasible on selected breeding material. 
Minor differences in berry sizes of only $1-2$~mm have to be achieved on thousands of grapevines at few days (ensure comparability of records), which is possible using image-based approaches. 
The framework presented in the current study aimed at an automated estimation of the size of grapevine berries from single images, which were taken in an experimental vineyard at different developmental stages. 
Hereby, the detection of representative berries and the determination of their diameter will be included.

The field experiments, obtained plant material and images are introduced in Section~\ref{sec:material} and Section~\ref{sec:images}. 
In Section~\ref{sec:framework} the proposed framework and its parts are introduced.
Section~\ref{sec:methods} explains the introduced parts in more detail.
The experiments and the obtained results are showed and discussed in Section~\ref{sec:experiments}.
The paper concludes in Section~\ref{sec:conclusion}.

\section{Material and Methods}
\subsection{Plant Material}
\label{sec:material}
Field experiments were conducted during the growing season of 2012. 
Tests involved rows of the Vitis vinifera ssp. vinifera cultivars 'Riesling', 'Pinot Blanc', 'Pinot Noir' and 'Dornfelder' at the experimental vineyard of Geilweilerhof located in Siebeldingen, Germany (N~49\degree 21.747, E~8\degree 04.678). 
Fifteen plants per cultivar were used for image acquisition and the measurement of reference data.

\subsection{Image acquisition and reference measurements}
\label{sec:images}
Image acquisitions were carried out using a single-lens reflex (SLR) camera (Canon\textsuperscript{\textregistered} EOS 60D). 
Camera calibration was performed according to \cite{Abraham1997} with a wide-angle of $28$~mm equivalent focal length. 
Images (8-bit RGB, $3456 \times 2304$  pixel) of grapevines were captured in the vineyard with a distance of about $1$~m at three different plant development stages BBCH~75, BBCH~81 and BBCH~89 (\cite{Bloesch2008}). 
The images were acquired under natural illumination field conditions with manually controlled exposure. 
Images were saved for offline processing. 
Reference measurements were conducted manually in parallel to image acquisition. 
Therefore, 50 berries per plant, cultivar and BBCH stage were randomly selected to measure the berry diameter by the utilization of an electronic calliper (Insize\textsuperscript{\textregistered} Co.LTD, Conrad electronics SE, Hirschau, Germany). 
In order to transform measurements in the images from pixel to mm, colored labels with a width of $13$~mm (Roth\textsuperscript{\textregistered} GmbH, Karlsruhe, Germany) were fixed on the wires in the vineyard.

\subsection{Framework}
\label{sec:framework}
\begin{figure}[ht]
	\centering
  	\framebox{\includegraphics[width=1\textwidth]{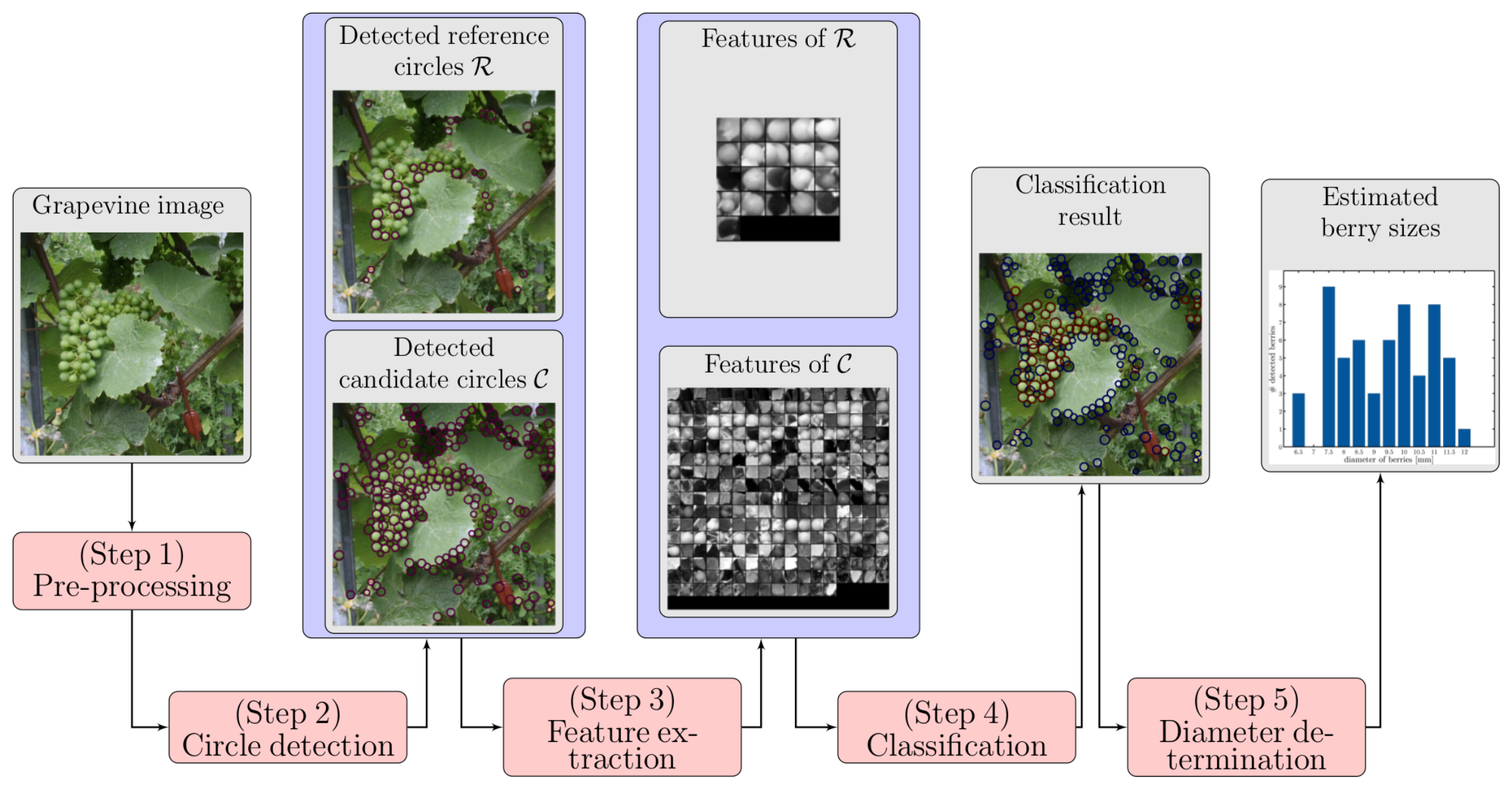}}
  	\caption{Image analysis framework for automated detection and measurement of grapevine berries. In (Step 1) the images are pre-preprocessed for (Step 2) in which circles for a reference set $\mathcal R$ and a candidate set $\mathcal C$ are detected. Various complementary features used for the classification are extracted in (Step 3). The detected candidate berries are classified in either 'berry' or 'non-berry' in (Step 4). In the end in (Step 5) the berry sizes are determined.}
\label{fig:framework}
\end{figure}

A five-step framework was developed using Matlab\textsuperscript{\textregistered} (Mathworks, Ismaning, Germany) in order to extract phenotypic data from images (Figure~\ref{fig:framework}).
The steps include various image analyzing tools and interpretation methods, which are explained in more detail in Section~\ref{sec:methods}.
The challenge of the framework is the detection of as many berries as possible in order to extract a representative amount of phenotypic data while keeping the error rate of falsely detected berries as low as possible in order to ensure a high quality of the extracted data.

\paragraph{(Step 1) Pre-processing} The image is adjusted automatically regarding brightness, color and contrast in order to compensate illumination effects.
	For this the image is converted into the YIQ color space and adjusted, whereas Y is the luminance and I and Q contain the chrominance information. Moreover, the contrast is stretched.
	
\paragraph{(Step 2) Detection of circular structures (see Section~\ref{sec:circles})} Two sets of circles are determined using circular Hough transform (\cite{Peng2007}):
		\begin{itemize}
			\item Automated detection of reference circles $\mathcal R$: Reference berries are image patches which are showing distinct circular structures. Assuming that the most dominant circles in one image are berries which can be used as training data in the classification process, the circle detector is applied with high constraints, \ie the detector returns only very distinct circles.
			\item Automated detection of berry candidates $\mathcal C$: Candidates for grapevine berries are all image patches which consist of at least a weak circular structure potentially showing a berry. The candidates are extracted by the circle detector using weak constraints, \ie the detector also returns circles with low responses. The reference set is a subset of the candidate set, whereas all candidates represent the test data for the classification process.
			The test data is classified into the class 'berry' and 'non-berry'.
		\end{itemize}
	
\paragraph{(Step 3) Feature extraction (see Section~\ref{sec:features})} Complementary features, namely color, histogram-of-oriented gradients (\cite{Dalal2005}) and gist (\cite{Oliva2001}), are extracted from image patches around the detected circles. The high-dimensional features are transformed into a low-dimensional feature space and used as the input for the classification process.

\paragraph{(Step 4) Classification of the image patches (see Section~\ref{sec:classification})}
The classification of the image patches is performed in two steps.
	 	\begin{itemize}
	 		\item Estimation of posterior probabilities: In order to estimate posterior probabilities, feature-wise thresholds are derived from the training data. 
	 		After the application of the thresholds to the test data, the output is transformed with a sigmoid function into probabilities.
	 		\item Application of a conditional random field (\cite{Lafferty2001}): A conditional random field is used to classify the extracted features of the candidates into the classes 'berry' and 'non-berry'. 
It uses the estimated posterior probabilities and prior knowledge about the spatial arrangement of berries, \ie that grapevine berries are arranged in clusters and have similar features such as color.
\end{itemize}
	 		
\paragraph{(Step 5) Determination of the berry size (see Section~\ref{sec:berrySize})} The size of the berries in the image is derived from the diameter of the detected circles classified as 'berry' and a single scale in order to transform pixel values into mm. 
A more accurate result can be derived by using a depth map, which assigns a depth to each pixel rather than one depth to all pixels.

\subsection{Image Analysis and Interpretation Methods}
\label{sec:methods}
In the following section the proposed berry detection framework is introduced in more detail.
Vectors $\d g = [g_i]=\left[g_1, \ldots, g_I\right] \trans$ are denoted with small bold symbols and matrices $\m G = [g_{ij}]=\left[\d g_1, \ldots, \d g_J\right]$ with elements $g_{ij}$ and column vectors $\d g_j$ with capital symbols. 
Calligraphy symbols are used for sets.
The elements (scalars or vectors) of a set $\mathcal G$ can be collected in a vector $\d g$ or a matrix $\m G$ by concatenation, using the same letter of the alphabet. 

\subsubsection{Circle Detection}
\label{sec:circles}
The circular Hough transform presented by \cite{Peng2007} is utilized with some modifications in order to detect circular shaped objects like berries in images and estimate their position~$\d z$ and radius~$\d d$.
The values of possible diameters are restricted to a range $\left[5\text{mm}~20\text{mm}\right]$.

\begin{figure}[ht]
  \centering
	\subfigure[Original image]{\includegraphics[height=0.2\textheight]{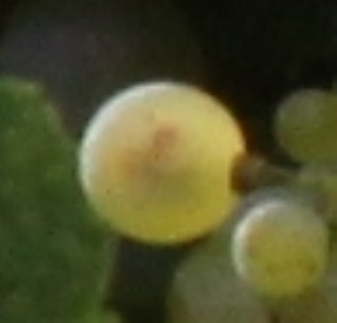}\label{fig:img}}\quad 
	\subfigure[Gradients]{\includegraphics[height=0.2\textheight]{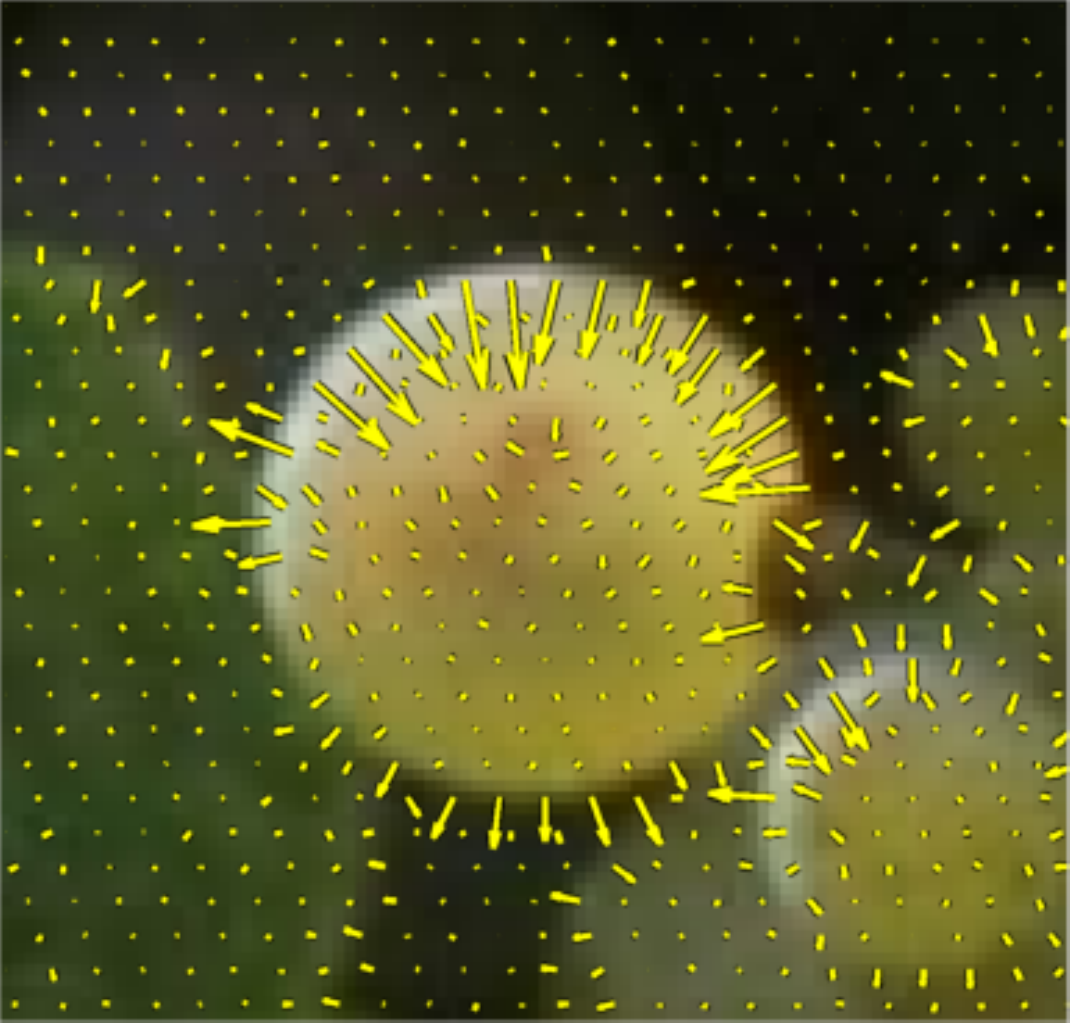}\label{fig:grad}}\quad  		
  		\subfigure[Accumulation array]{\includegraphics[height=0.2\textheight]{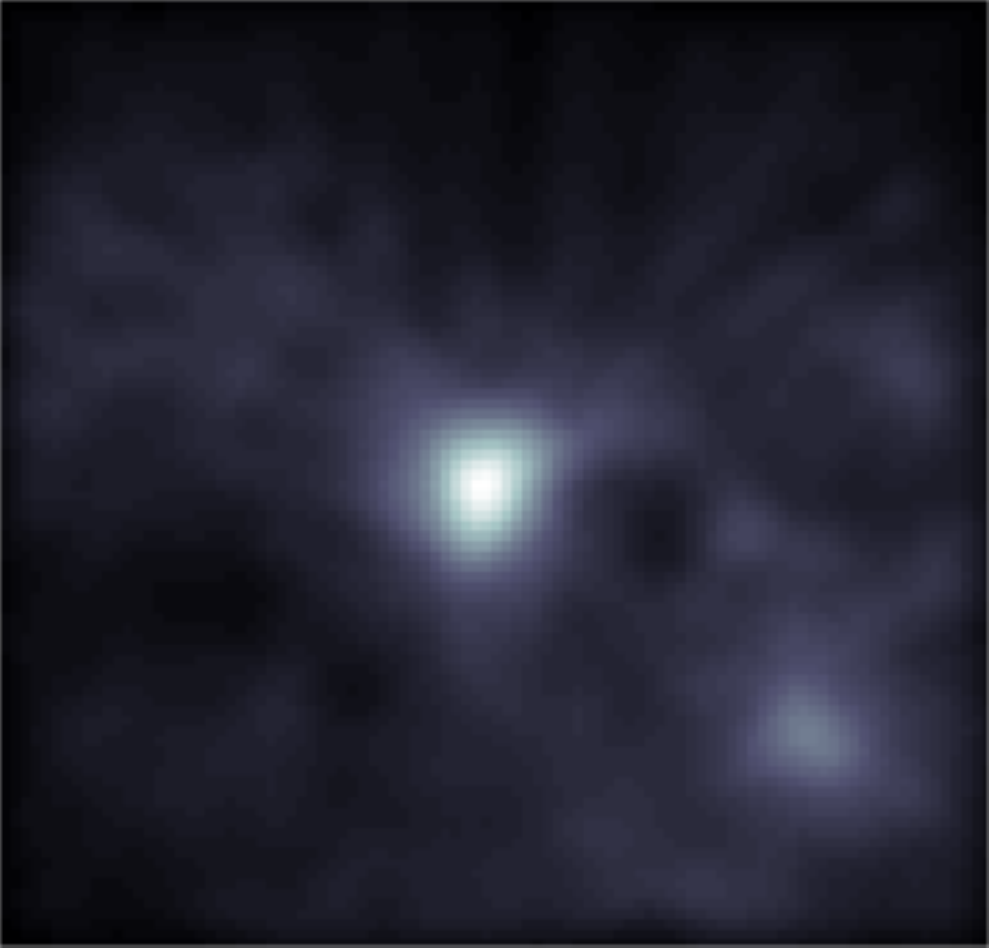}\label{fig:peaks}}\quad
  		\subfigure[Signature curve]{\includegraphics[height=0.2\textheight]{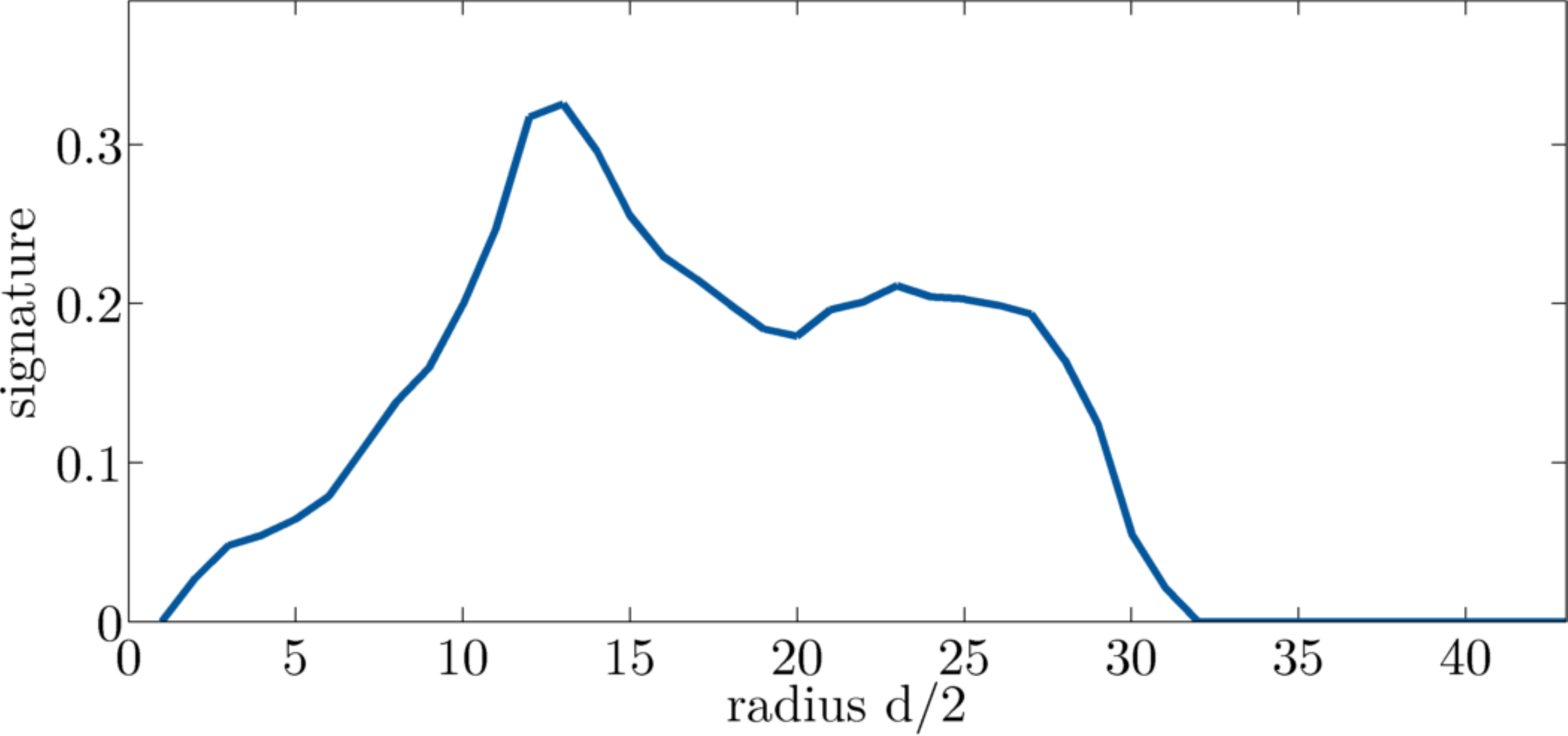}\label{fig:signature}}\quad
      \subfigure[Detected Circle]{\includegraphics[height=0.2\textheight]{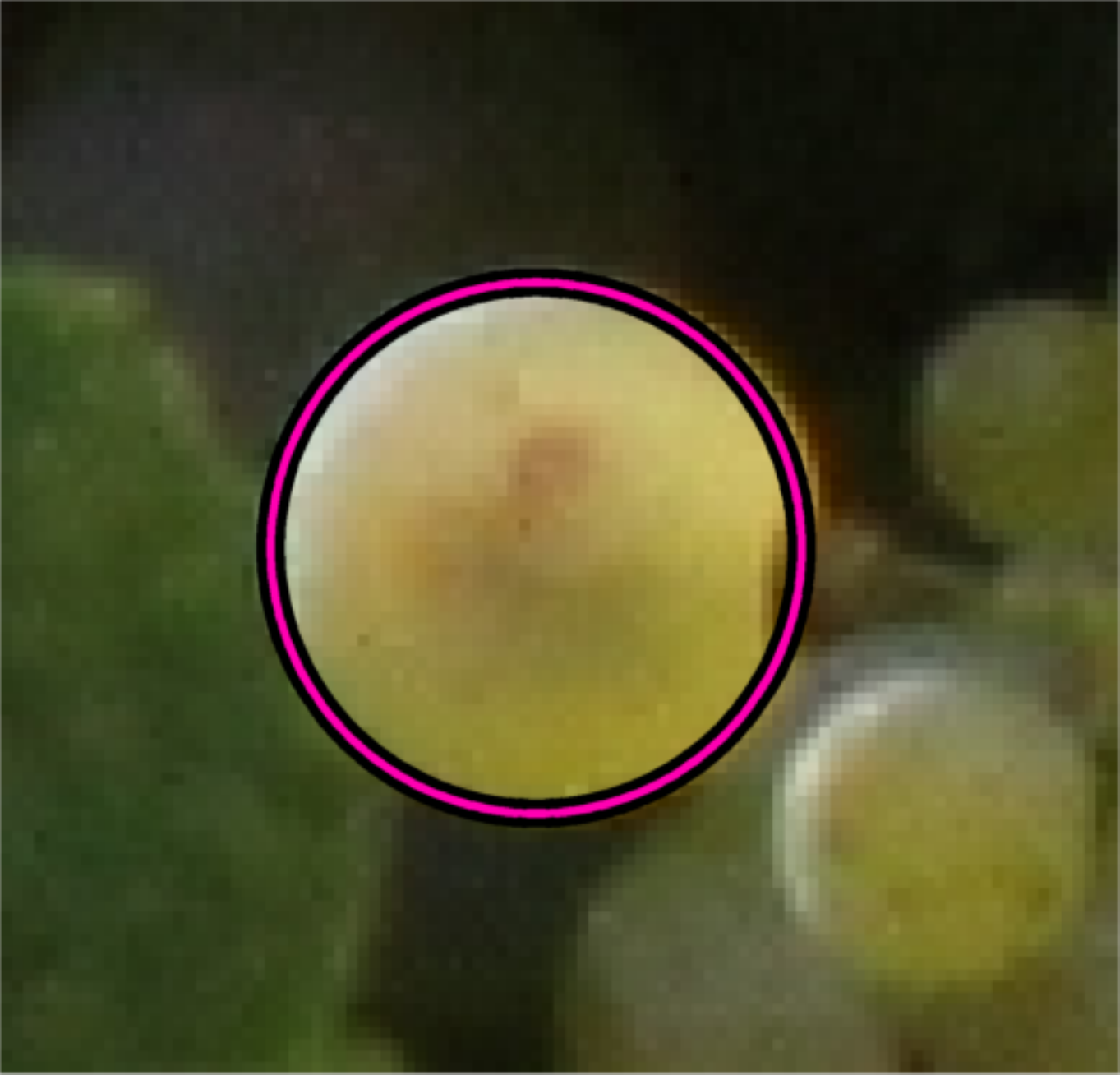}\label{fig:detected}}
  \caption{Illustrated steps of the detection of circular structures. From the original image the gradients are obtained using a Sobel filter. The gradients are transformed to an accumulation array, whereas the bright peak indicates the position of the circle center. Using the signature curve the radius of the circle can be obtained.}
  \label{fig:cht}
\end{figure}

First, a Sobel filter is applied to the $M\times N$-dimensional gray-valued image $\m I$ yielding the gradient image in vertical direction $\m I_u$ and in horizontal direction $\m I_v$.
The magnitude is given by $\m G~=~\sqrt{\m I_u^2 + \m I_v^2}$. 

In order to find circular structures, the gradient field is converted to an accumulation array $\m A$ of the same size.
For this $\m G$ is first thresholded by $t_g$ yielding the binary image $\widehat{\m G}$.
In a second step, each non-zero element in $\widehat{\m G}$ votes for several positions in the accumulation array with weights $\d w$. 
High values in the accumulation array are indicating centers of circles.

Contrary to \cite{Peng2007} in this framework the threshold is determined automatically utilizing a standard deviation ridge detector (\cite{Hidayat2009}).
The output of the standard deviation ridge detector is denoted with $\m S$, whereas the pixel-wise values of $\m S$ are the standard deviation calculated in the local neighborhood of each pixel in $\m I$. 
The largest values are indicating boundaries between regions with different textures.
The threshold $t_g$ is determined by testing several values for $t_g$ and correlating the obtained binary $\widehat{\m G}$ with $\m S$.
The largest correlation coefficient indicates the best threshold. 
In this way, enough gradients are suppressed in order to remove noise and distinct gradients remain in order to search for circular structures.
Using a fixed value for the threshold in the proposed framework would not lead to good results due to changing image conditions such as illumination and berry color and thus, changing magnitudes of the gradients. 

Each non-zero value in $\widehat{\m G}$ votes for several positions in the accumulation array, namely for all coordinates of pixels that lie on the line segment defined by the gradient direction and the range of possible radii.
Because the gradient directions point either towards the circle center or away from it, the sign of the gradient is omitted and the vote is added in both directions. 
For each vote the weights $\d w$ are derived from the output of the standard deviation ridge detector $\m S$.
The votes are accumulated and peaks in the accumulation array indicates probable positions of circle centers, as can be seen in Figure~\ref{fig:cht}.
In order to find distinct peaks, the array is smoothed and a local maximum filter is applied. 
Moreover, the array is thresholded by $t_{a,r} = 0.6\max{(\m A)}$ yielding reference circles and $t_{a,c} = 0.3\max{(\m A)}$ yielding candidate circles.

The radii of the detected circle centers are determined using so-called signature curves (Figure~\ref{fig:signature}).
A signature curve belonging to a detected circle center is a function of the radius.
The function value of the curve is computed from the gradients supporting a circle when choosing a certain radius. 
The more distinct the circular structure given a specific radius is, the higher is the response.
A more detailed description of the signature curves can be found in \cite{Peng2007}.

\begin{figure}[ht]
  \centering
	\subfigure[Detected circles without the refinement]{\includegraphics[width=0.4\textwidth]{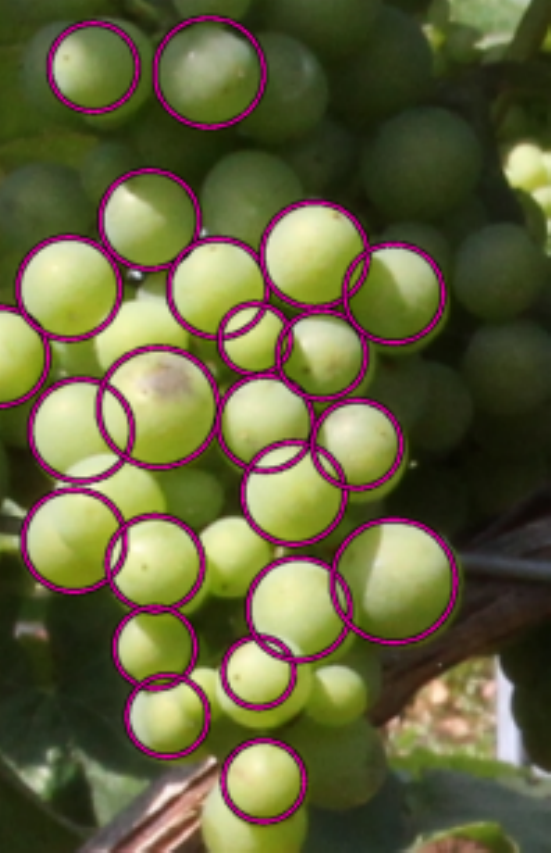}}\qquad 
	\subfigure[Detected circles after the refinement]{\includegraphics[width=0.4\textwidth]{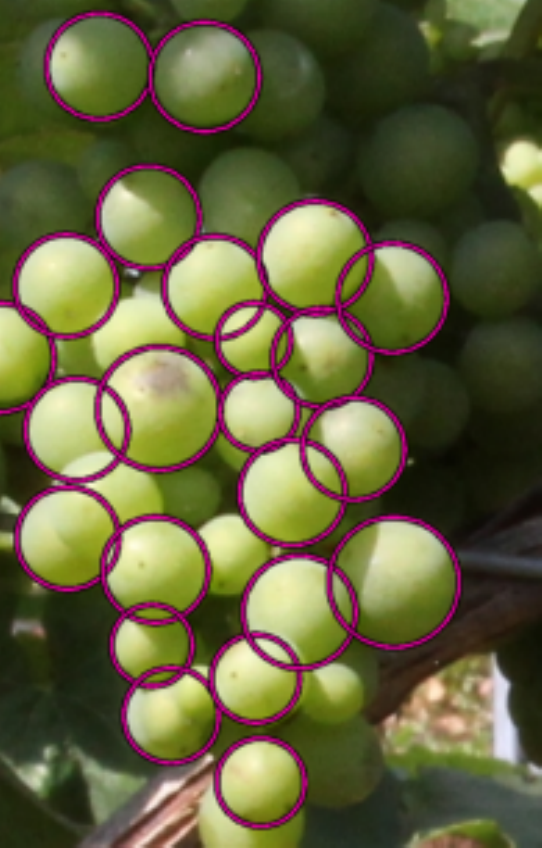}}
  \caption{The circle centers and radii of the detected circles (left image) are refined using a sliding window approach. The detected circles in the right image fit better to the border of the grapevine berries than the output of the circular Hough transform (left image).}
  \label{fig:adjusted}
\end{figure}

Since the images are very cluttered and the gradient directions can be noisy, contrary to \cite{Peng2007} another step is introduced in order to refine the positions and radii of the detected circles.
For this a sliding window approach is used, which is a localized search over space and scale.
In this framework, for each detected circle a small accumulation array is built with three dimensions: shift of circle center $\d z$ in vertical direction, shift of circle center $\d z$ in horizontal direction, scale of the radius $d$. 
Based on the current position and radius of a circle the position of the circle center is shifted in both directions within a range of $[-5\text{pixel} ~ 5\text{pixel}]$ and the radius is scaled within a range $[-5\text{pixel}~5\text{pixel}]$ under the condition that the adjusted radius must lie within the restricted range introduced in the first paragraph in this section.
For each shift and scale a circle is constructed with discrete pixel coordinates.
The sum of the pixel values in the image $\m S$ with these coordinates are used as weight in the accumulation array.
The peak in the accumulation array indicate the shift and scale of the detected circle.
One example can be seen in Figure~\ref{fig:adjusted}.

Summarizing, the circle detector uses a coarse-to-fine strategy, since first the positions and radii of the circles are roughly determined using a circular Hough transform and afterwards adjusted using a sliding window approach.

The detected circle centers and radii of the reference circles and candidates are collected in the reference set $\{\d z_r, d_r\} \in \mathcal R$, $r=1,\ldots,R$, and candidate set $\{\d z_c, d_c\} \in \mathcal C$, $c=1,\ldots, C$, respectively.
The reference set is a subset of the candidate set $\mathcal R \in \mathcal C$.

From the detected circles the training and test data for the classification is derived.
All candidates are meant to be classified and are thus the test data. 
Training data is necessary to learn a classification model from which each candidate can be classified. Since there is no training data given in advance, an active learning approach is used. 
This approach uses the assumption that most of the detected reference circles are berries and thus, can be labeled as the class 'berry'.	
To be robust against sporadic falsely detected reference circles, not all reference circles are used as training data, which will be explained in more detail in Section~\ref{sec:classification}.
The training data is actively acquired from scratch by the circle detector for each new image and thus, the classification model automatically adapts to changing conditions such as illumination or berry color.

\subsubsection{Feature Extraction}
\label{sec:features}
Within the framework color features $\d x_{\text{rgb}, c}$, gradient based histogram-of-oriented-gradients (HoG) features $\d x_{\text{hog}, c}$ as well as gist features $\d x_{\text{gist}, c}$ describing the dominant spatial structure are used.
Due to the separate treatment of these highly complementary features, they can be used with different weights in the conditional random field ensuring a best possible discrimination of patches into the classes 'berry' or 'non-berry'.

Quadratic image patches of the size $\left(2d_c \times 2d_c\right)$ around the circle centers are defined, whereas $d_c$ is the radius of the $c$-th detected circle.
All patches are resized to an uniform size of $N_p\times N_p$ to ensure an equal feature dimension for the candidates.
While the dimension of HoG and gist features are independent of the patchsize, the resizing is necessary when extracting color features, which are used vectorized.

\paragraph{Color Features}
Although the berries do not significantly differ from leafs and grass, which makes most of the background, the features can be used to discriminate circles which are berries and these which are positioned on canes, artificial background objects and ground. 
RGB color features are extracted and vectorized, so that a $(N_p\cdot N_p\cdot 3)$-dimensional feature vector $\d x_{\text{rgb}, c}$ is assigned to each candidate.

\paragraph{Histogram-of-Oriented-Gradients Features}
Besides the color features also gradient information are used, which are represented as histogram-of-orientated-gradients (HoG) features (\cite{Dalal2005}).
The structure of an image patch is described by the distribution of the magnitude and directions of gradients, in which the circular structure of a berry yield a characteristic HoG descriptor.
Here, the image patches are convolved with a Sobel filter and divided into a fixed number of quadratic regions of equal size.
In each region a orientation-based histogram of the unsigned gradients directions comprising a fixed number of bins is computed, whereas each gradient value casts a weighted vote for the histogram.
The histogram entries are concatenated and vectorized yielding a feature vector $\d x_{\text{hog}, c}$ assigned to each candidate.

\paragraph{Gist Features}
Gist features are used in order to represent the dominant spatial structure of a patch/image such as roughness or openness.
Following \cite{Oliva2001} a gist descriptor is built based on a very low dimensional representation of the scene called spatial envelope.
In order to describe the patch, a discrete Fourier transform is performed yielding the amplitude spectrum of the gray-valued image.
The amplitude spectrum gives information about the structure of the image, such as the orientation and smoothness of object contours.
Additionally, the energy spectrum is derived, which is the squared magnitude of the amplitude spectrum.
Instead of using the Fourier transform for the whole image, a windowed Fourier transform for uniformly arranged, overlapping image parts is applied.
Based on the values of the energy spectrum for each part the gist descriptor is built yielding a feature vector $\d x_{\text{gist}, n}$ assigned to each candidate.

\paragraph{Transformation of the Features}
One challenge for one-class classification is the choice of the features for the target class and associated with this the definition of a suitable decision boundary in order to distinguish  the target objects from all other objects.
Therefore, instead of using the features in the original feature space, they are transformed into a lower-dimensional space. 
This makes it easier to find thresholds defining the decision boundary.
From the decision boundaries posterior probabilities for each candidate can be derived.

The feature transformation follows the idea of correlation coefficient clustering proposed by \cite{Hsu2010}, in which data points with similar features are grouped in clusters when using their mutual correlation coefficients.
Since it can be assumed that berries in one image have similar features, the new features $l_{(\cdot),c}$ of the candidates are derived from the median correlation to the reference patches
\begin{equation}
	l_{(\cdot),c} = \operatorname{med}_r\(\rho\(\d x_{(\cdot),c}, \d x_{(\cdot),r}\)\)\,,
\end{equation}
where $\rho\(\d x_{(\cdot),c}, \d x_{(\cdot),r}\)$ is the correlation coefficient between the candidate feature vector $\d x_{(\cdot),c}$ and the reference feature vector $\d x_{(\cdot),r}$.
The kind of feature is generalized denoted with $(\cdot)$. 
Then small values indicate a faint resemblance to the reference patches and high values indicate a close resemblance, \ie candidates with high feature values are most probable berries.
The median is used in order to robustly define a threshold, which should meet the following conditions for the purpose of finding a representative amount of berries: circles with correlations coefficients larger than the threshold are most certainly berries and circles with correlations coefficients smaller than the threshold are no berries or precarious berries.

The introduced rgb, HoG and gist features are treated separately in the classification process for the feature-wise assignment of weights in the conditional random field depending on their discriminative power. However, an alternative representation as a concatenated feature vector is also possible.

\subsubsection{Classification}
\label{sec:classification}
Since not all candidates which are detected by the circle detector are berries, a classification problem is formulated in order to classify each candidate into the target class 'berry' or all other objects belonging to the class 'non-berry'.
The classification is done via a conditional random field, which uses posterior probabilities of the features as well as prior knowledge of the spatial relations between candidates.

\paragraph{Estimation of Posterior Probabilities of the Test data}
\label{sec:probabilities}
In order to derive posterior probabilities for the target class 'berry' and all other objects denoted with 'non-berry', two steps must be performed.
First, for each kind of feature a threshold must be found from which each candidate based on its feature vector can be assigned to one class with a certain degree of confidence. 
Second, the confidence is transformed into probabilities.

The threshold $t_{(\cdot)}$ for a feature~$(\cdot)$ is chosen to be the $p$-percentile of the set of all correlation coefficients between the reference circles $\{\rho\(\d x_{(\cdot),r}, \d x_{(\cdot),r'}\)\}$ with $r\neq r'$.
The parameter $p$ is the value so that $p$ percent of all correlation coefficients are smaller than $t_{(\cdot)}$.
The larger $p$ is, the more candidates will be classified as 'berry'. 
The percentile is used in order to be robust against noise, effects such as occlusions, clutter and illumination changes and incorrectly classified reference circles.

This can be formulated in a probabilistic way by stating that all candidates whose feature vector $l_{(\cdot),c}$ is smaller than the threshold $t_{(\cdot)}$ get a probability smaller than $0.5$ and are unlikely 'berry'.
The probabilities can be obtained by a sigmoid transformation
\begin{align}
	P\(y_c = \text{'non-berry'}| \d x_{(\cdot),c}\) & = \frac{1}{\( 1 +\exp\(s\( l_{(\cdot),c} - t_{(\cdot)} \) \) \)}\,,\\
	P\(y_c = \text{'berry'}| \d x_{(\cdot),c}\) & = 1- P\(y_c = \text{'non-berry'}| \d x_{(\cdot),c}\)\,,
\end{align}
with $s$ defining the sharpness of the probabilities.

\paragraph{Conditional Random Field}
\label{sec:crf}
A conditional random field (CRF, \cite{Lafferty2001}) is an undirected graphical model, which is used to incorporate prior knowledge about the spatial relations between the candidates. 
Because neighbored candidates tend to have the same class label ('berry' or 'non-berry') if their features closely resemble each other, an irregular graph structure is introduced, which models the connection between these candidates. 
Besides this and the posterior probabilities in Section~\ref{sec:probabilities} also the average distance to neighbors are used within the model in order to prevent that isolated circles are classified as 'berry'.

The conditional random field model is defined as
\begin{align}
	E(\vlabels) & =
    - w_{\text{rgb}} \sum_{c} \log P\(y_c| \d x_{\text{rgb},c}\)
    - w_{\text{hog}} \sum_{c} \log P\(y_c| \d x_{\text{hog},c}\)\nonumber\\&\hspace{1.2em}
    - w_{\text{gist}} \sum_{c} \log P\(y_c| \d x_{\text{gist},c}\)
    - w_{\text{dist}} \sum_{c} \log P\(y_c| \d x_{\text{dist},c}\) \nonumber\\&\hspace{1.2em}
    + w_{\text{spatial}} \sum_{(c,c') \in \mathcal N} \Phi\(y_c,y_{c'}, \d x_{\text{con},c}, \d x_{\text{con},c'}\)\,.
    \label{eq:crf}
\end{align}
The class labels of the circles are given by $y_c$, which are either 'berry' or 'non-berry'.
The first three, unary terms are defined as the negative logarithms of the posteriors described in Section~\ref{sec:probabilities}. 
The fourth, unary term is the negative logarithm comprising the average distance to neighbored circles.
Since the final labeling of the candidates is assumed to be smooth within the
image, \ie neighbored candidates have the same class label, this prior knowledge is introduced by means of a data-depended Potts model in the fifth, binary term.
The variable $\d x_{\text{con},c}$ is the concatenation of all features which are used in the unary terms, whereas $\d x_{\text{dist},c}$ are transformed to a range $[-1~1]$.
The set of spatial neighbors is denoted by $\mathcal N$.
The variables $w_{(\cdot)}$ are the weights between the terms.
Graph-cut (\cite{Boykov2001}) is used to solve for the best labeling
$\widetilde{\vlabels} = \operatorname{argmin}_{\vlabels}
E(\vlabels)$.

Besides the posterior probabilities described in Section~\ref{sec:probabilities} also the average distance to neighbored candidates is used. 
Because berries are grouped in clusters it is more likely that isolated circles belong to the class 'non-berry'.
Therefore, the fourth, unary term is introduced that models the probability of a circle belonging to the class 'berry' or 'non-berry'.
From the neighborhood $\mathcal N$ the mean distance $l_{\text{dist},c}$ from each candidate to its neighbors can be derived with
\begin{equation}
	l_{\text{dist},c} = \frac{1}{N_c}\sum_{c'\in \mathcal N} \sqrt{\(\d z_c - \d z_{c'}\)\trans\(\d z_c - \d z_{c'}\)}\,,
\end{equation}
where $N_c$ is the number of neighbors.
The probability that a candidate belongs to one of the classes is given by
\begin{align}
P\(y_c = \text{'non-berry'}| \d x_{\text{dist},c}\) & = \max\(\frac{1}{\( 1 +\exp\(s\( l_{\text{dist},c} - t_{\text{dist}} \) \) \)}, ~0.5\)\,,\\
	P\(y_c = \text{'berry'}| \d x_{\text{dist},c}\) & = 1- P\(y_c = \text{'non-berry'}| \d x_{(\text{dist},c}\)\,.
\end{align}
	
The value $t_{\text{dist}}$ is set to 3 times the current median diameter of the reference circles.
The sigmoid function is cropped, because isolated candidates with no nearby neighbors are probable no berries, but vice versa candidates positioned nearby are not necessarily berries.
Their probability are set to $0.5$, so that the other terms decide whether these candidates are 'berry' or 'non-berry'.

The binary term is introduced in order to favor that neighbored circles with similar features get the same class label.
For example, if features of patches showing berries are uncertain regarding their class label (\eg caused by illumination effects or occlusions) and their neighbors are berries with similar features, the binary term guides the decision into the correct direction to classify the circle as 'berry'.

\begin{figure}[ht]
  \centering
  		\subfigure[Voronoi diagram for the derivation of the graph structure]{\includegraphics[width=0.45\textwidth]{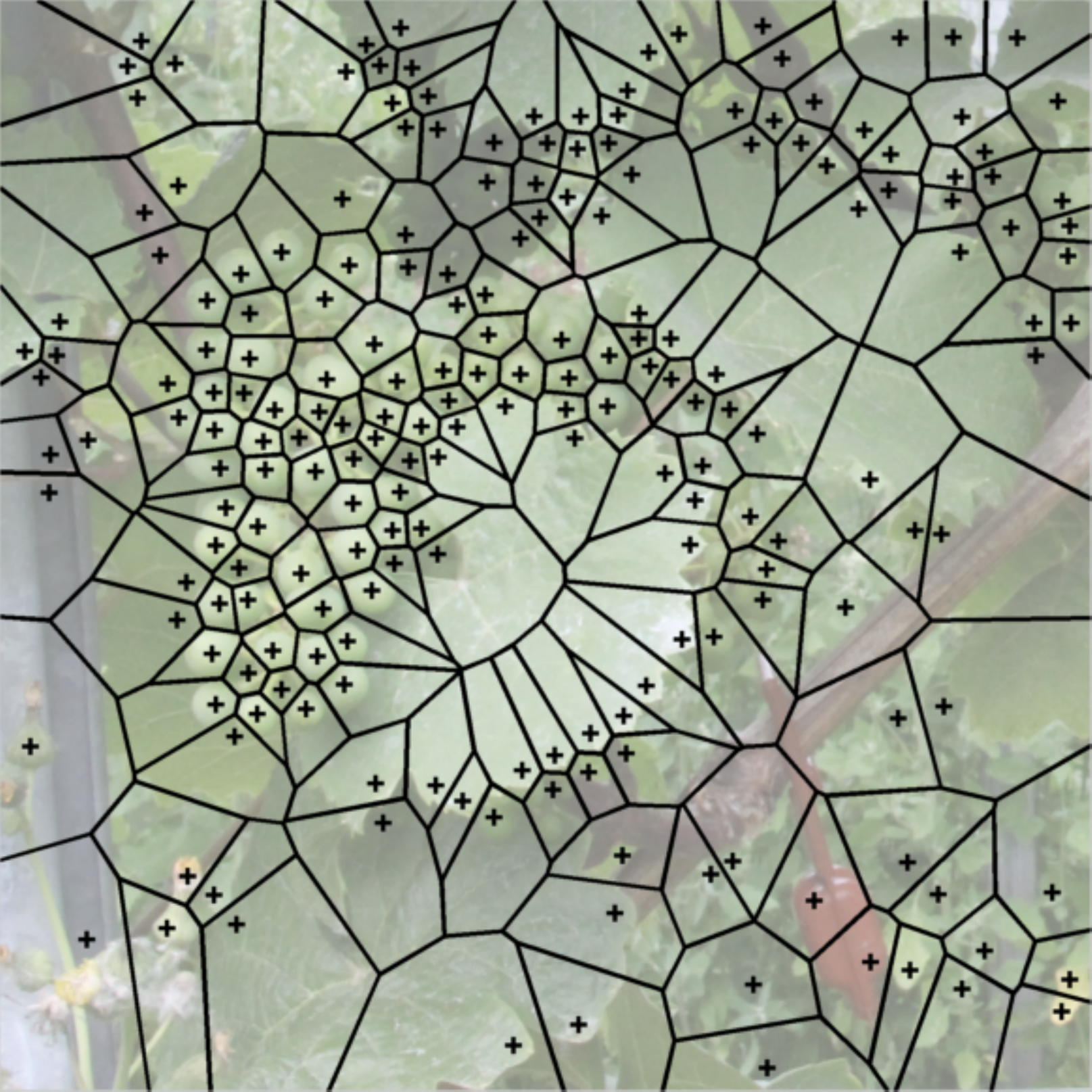}
  		\label{fig:voronoiDiag}}\quad
      \subfigure[Irregular graph structure]{\includegraphics[width=0.45\textwidth]{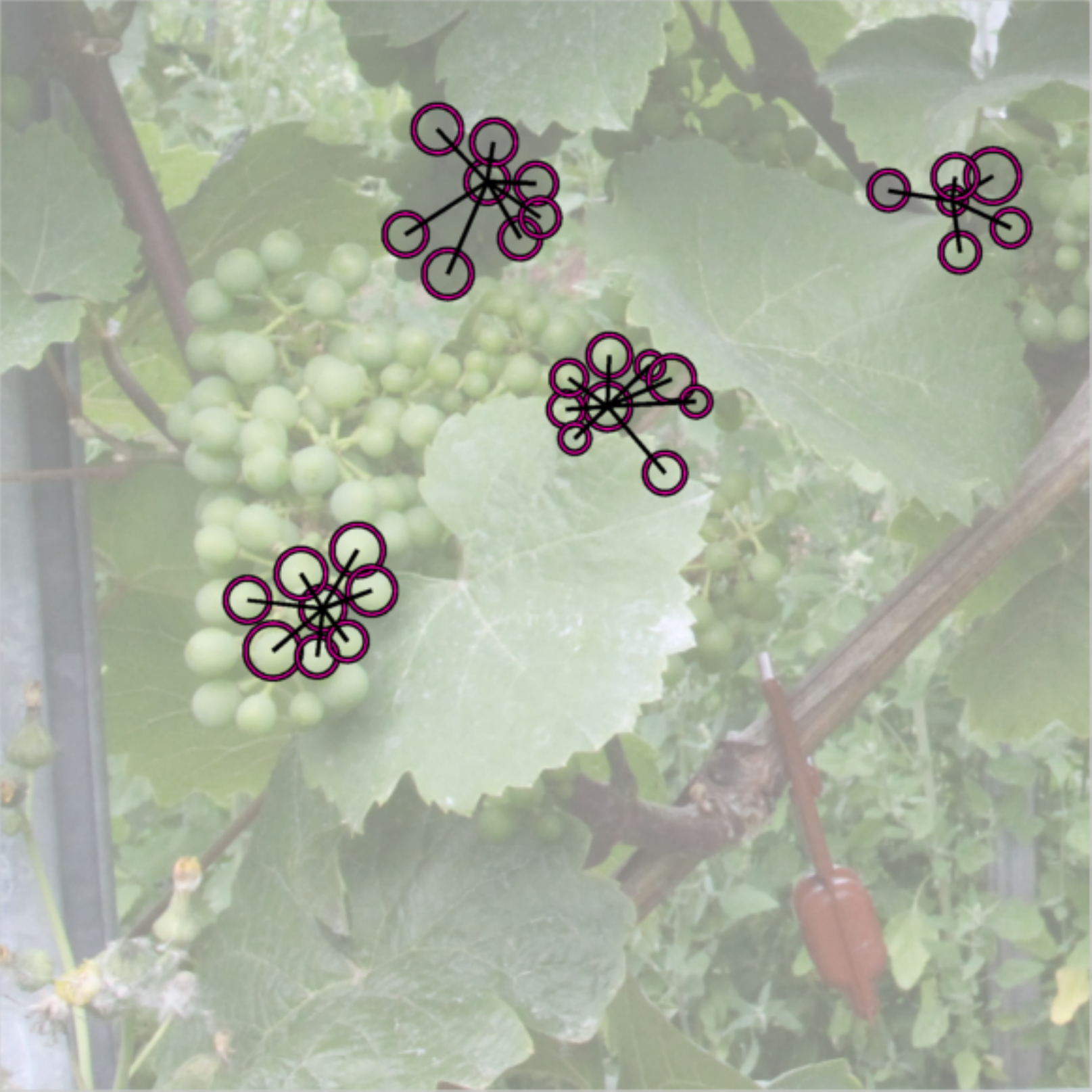}
        \label{fig:connections}}
  \caption{
    Spatial relations between the candidates. Each candidate is connected to its neighbors (right image) derived from neighbored cells in the Voronoi diagram (left image). As an example, the right image shows four candidates and its connections as black lines. All spatial relations define the irregular grid structure within the conditional random field.}
  \label{fig:crf}
\end{figure}

In order to define the neighbors of each candidate, an irregular graph structure is defined by a Voronoi diagram, whereas the positions $\d z_c$ of the circles are the centers of the diagram (see Figure~\ref{fig:voronoiDiag}).
Thus, adjacent cells indicate neighbored candidates.
Figure~\ref{fig:connections} shows four candidates and their neighbors, which where obtained from the Voronoi diagram.

The binary term is modeled as the euclidean distance between the concatenated features of two neighbored circles
\begin{equation}
	\phi_{cc'} = \sqrt{\(\d x_{\text{con},c} - \d x_{\text{con},c'}\) \cdot \(\d x_{\text{con},c} - \d x_{\text{con},c'}\)}\,,
\end{equation}
involving the dot product of both vectors. 
The concatenation of all features is defined as $\d x_{\text{con},c} = \left[\d x_{\text{rgb}, c}\trans, \d x_{\text{hog}, c}\trans, \d x_{\text{gist}, c}\trans, q/10~\d x_{\text{dist}, c}\trans\right]\trans$, where $q$ is the median distance between the candidates in order to scale $\d x_{\text{dist}, c}$ to a similar range as the other features.
The term is only considered if two neighbors $c$ and $c'$ get the same label, \ie if two neighboring candidates have a close resemblance of their features they are likely to have the same class label, but if they have a faint resemblance of their features it does not automatically indicate that their class labels are unequal.

In the following the set of candidates which are labeled with 'berry' is denoted as $\mathcal B$. 

\subsubsection{Determination of the berry size}
\label{sec:berrySize}
The detected circle diameters are obtained in pixel.
In order to transform the berry diameters into mm, a colored label of $13$~mm width was fixed at the steel wire, and was used to calculate the conversion ratio $a$ between mm and pixel.
The ratio is given by $a = \frac{13~\text{[mm]}}{b~\text{[pixel]}}$, where $b$~[pixel] is the width of the colored marking observed in the image.
Then the radius $d$~[pixel] of a circle in the image can be transferred to \mbox{$d'~\text{[mm]}=a\cdot d~\text{[pixel]}$}.
In these experiments the colored label is measured manually in the image. 
However, this process can by automated, \eg if a stereo camera system is used with a fixed and known basis.

\section{Experiments and Results}
\label{sec:experiments}

\subsection{Experimental Setup}
The experiments are conducted with the proposed framework, which is written in Matlab\textsuperscript{\textregistered}.
The images are resized to $1667\times2500$ and the luminance and chrominance as well as the contrast are adjusted using a Matlab\textsuperscript{\textregistered} file package available online\footnote{\url{http://www.mathworks.com/matlabcentral/fileexchange/24290-auto-enhancement-for-images}}.
The extraction of HoG features is done by an own implementation in Matlab\textsuperscript{\textregistered}.
There are $4$ cells used and $8$ orientations of unsigned gradients in each cell.
In order to extract the gist features the implementation of \cite{Oliva2001} is used, which is available for download\footnote{\url{http://people.csail.mit.edu/torralba/code/spatialenvelope/}}.
The number of cells for the windowed Fourier transform is set to $4\times 4$, the number of scales is chosen to be $4$ and the number of orientations is set to $8$ for each scale.
For the conditional random field the parameters are derived experimentally and set to $w_{\text{rgb}}=0.5$, $w_{\text{hog}}=0.5$, $w_{\text{gist}}=1$, $w_{\text{dist}}=2$ and $w_{\text{spatial}}=0.5\left(w_{\text{rgb}}+w_{\text{hog}}+w_{\text{gist}}+w_{\text{dist}}\right)$.
The weights are chosen best according to the discriminative power of the feature.
Alternatively, the parameters can be learned using for example maximum likelihood, see \cite{Korvc2008}.
The value for $p$ for the definition of the percentile is set to $50$, which represents the median value.
If no circles were classified as 'berry', the value was increased to $0.7$.

For the evaluation the mean diameter and the standard deviation are compared to the manual reference measurements.
All estimated diameters are rounded to $0.5$~mm steps and represented in a histogram.
From the histogram the occurrence and frequency of the estimated diameters can directly be derived.

\subsection{Results and Discussion}
Three important berry development stages of grapevine were investigated 1)~BBCH~75 -- the pea size of berry development; 2)~BBCH~81 -- the beginning of ripening when berries start softening; and 3)~BBCH~89 -- the end of ripening and time of harvest.
Table \ref{tab:results} shows the manual results and the obtained results of the framework for these data sets. 

\renewcommand{\baselinestretch}{1}
\begin{table}[ht]
\centering
\caption{Sizes of three developmental stages of grapevine berries: 1)~BBCH~75 -- the pea size of berry development; 2)~BBCH~81 -- the beginning of ripening when berries start softening; and 3)~BBCH~89 -- the end of ripening and time of harvest. Reported are the number of detected berries \# $\mathcal B$, the manual measurement of the mean diameter with the standard deviation in brackets, the estimated mean diameter and standard deviation in brackets obtained by the framework as well as the mean differences between the manual measurements and the estimated diameters (MD) and the mean absolute differences between the manual measurements and the estimated diameters (MAD).}
\begin{tabular}{llcccccc}
\toprule
Stage & Sort & Mean \#$\mathcal B$ & \multicolumn{4}{c}{Mean diameter [mm]}\\
\cmidrule(rl){4-7}
& & & Manually & Framework & MD & MAD\\
\midrule
BBCH~75  &  Riesling      		& 71.1 & 8.5 (1.1) & 9.5 (0.7) & 0.9 & 1.0\\
		 &  Pinot Blanc 		& 67.1 & 8.6 (1.2) & 8.7 (0.8) & -0.1  & 0.7\\
		 &  Pinot Noir 			& 63.4 & 7.8 (1.4) & 9.8 (1.0) & 2.0 & 2.0\\
		 &  Dornfelder    		& 83.1 & 8.9 (1.1) & 10.1 (0.5) & 1.2 & 1.2\\
\cmidrule(rl){1-7}		 
BBCH~81  &  Riesling      		& 156.7 & 11.8 (1.2) & 11.8 (0.7) & 0.0 & 0.6\\
	     &  Pinot Blanc 		& 75.1 & 11.9 (1.2) & 12.4 (0.4) & 0.6 & 0.6\\
         &  Pinot Noir 			& 148.9 & 10.1 (1.6) & 12.3 (1.0) & 2.1 & 2.1\\
		 &  Dornfelder    		& 202.7 & 12.1 (1.1) & 13.2 (0.8) & 1.1 & 1.2\\
\cmidrule(rl){1-7}		 	
BBCH~89  &  Riesling      		& 102.7 & 13.5 (1.3) & 14.2 (1.2) & 1.1 & 1.5\\
	     &  Pinot Blanc 		& 232.2 & 13.5 (1.2) & 13.9 (0.6) & 0.4 & 0.5\\
         &  Pinot Noir 			& 90.0 & 11.8 (1.8) & 13.4 (0.9) & 1.6 & 1.7\\
		 &  Dornfelder    		& 112.7 & 15.4 (1.4) & 15.2 (1.5) & -0.1 & 1.0\\		 
		 	
 \bottomrule
\end{tabular}
\label{tab:results}
\end{table}
\renewcommand{\baselinestretch}{2}

\begin{figure}[ht]
  \centering
	\includegraphics[width=1\textwidth]{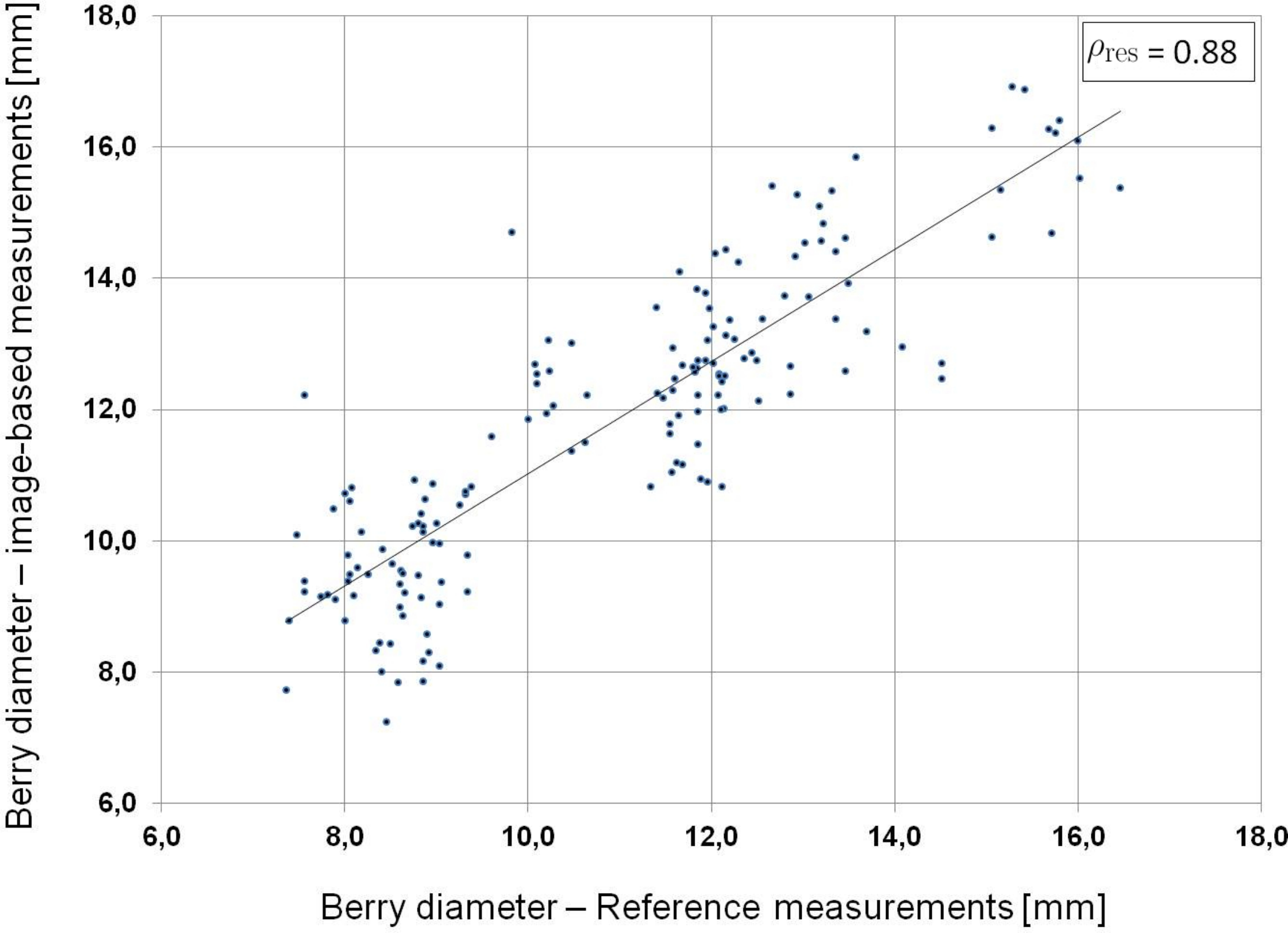}
	\label{fig:voronoi}
  \caption{Correlation plot of the manually measured and image-based measured berry diameter per image ($\rho_{\text{res}} = 0.88$)}
  \label{fig:correlation}
\end{figure}

The average difference for all images showing berries in BBCH~75 is $1.0$~mm, for
BBCH~81 $0.9$~mm and for BBCH~89 $0.6$~mm and the average absolute difference for all images showing berries in BBCH~75 is $1.2$~mm, for BBCH~81 $1.1$~mm and for BBCH~89 $1.2$~mm.
The average difference for 'Riesling' computed for all stages of growth is $0.7$~mm, for 'Pinot Blanc' $0.3$~mm, for 'Pinot Noir' $1.9$~mm and for 'Dornfelder' $1.1$~mm and the average absolute difference for 'Riesling' computed for all stages of growth is $1.0$~mm, for 'Pinot Blanc' $0.6$~mm, for 'Pinot Noir' $1.9$~mm and for 'Dornfelder' $0.7$~mm. 
Thus, the obtained averages (absolute) differences are similar when comparing the growth of stages, but vary for several sorts. 
The mean berry size of a plant estimated by the framework correlates with the manual measurements by $0.88$, as illustrated in Figure~\ref{fig:correlation}.

There is an overestimation in the berry's diameter for nearly all data sets. 
The main reason is that berries with a large diameter generally are more likely to be detected than berries with small diameters since their structure is more distinct.
Contrary to this, the manual reference measurements were randomly selected.
This could be an explanation for the surpassing overestimation for 'Pinot Noir' of $2.0$~mm. 
In 2012 the berry sizes per grape cluster especially varying for 'Pinot Noir' in contrast to 'Riesling', 'Pinot Blanc' and 'Dornfelder' underlined by the standard deviation over the mean diameters given in brackets in Table~\ref{tab:results}.   
Therefore, taking the mean of the estimates obtained from the framework lead to an overestimation since larger diameter have a higher weight due to their frequent occurrence in the detection result. 
Factors influencing this effect are the variations of the berry sizes, the compactness and arrangement or the color of the berries.
Thus, a histogram of diameters is more meaningful than only using the mean in order to interpret the results. 

In general, OIV descriptors are applied in order to classify grapevine traits, whereas the berry diameter is estimated using the OIV descriptor number 221. 
In contrast to the proposed image-based framework, the application of the OIV descriptor classifies the berry size into only five classes covering all expected sizes from $<8$~mm up to $>28$~mm. 
Thus, the usage of this classification resulting in missing precision.
Moreover, beside the necessity that experts are needed, the results of the visual estimated berry diameter by humans are subjective resulting in error variations between the results of different people.
Nevertheless, precise berry size data ($1-2$~mm accuracy) are required from a mapping population of several hundreds of individual plants in order to enable fine mapping of QTL regions or in order to determine the berry size of cultivars before harvest. 
The non-invasive image capture of plants in the field followed by the automated image analysis framework ensures a more comprehensive phenotypic analysis in a high-throughput manner. 
It also enables phenotypic evaluation from several plants per genotype/cultivar which ensures several biological repeats.

Moreover, it should be noted that the manual estimation of a sufficient amount is very time consuming and usually it is not feasible within the regularly breeding programs. 
In the conducted experiments the manual measurement of $50$ reference berries directly in the field by using a digital caliper needs $4$~minutes per plant. 
Thus, precise berry size data could be recorded by hand from $15$ grapevines in one hour. 
In comparison to that, $10$~seconds are needed in order to capture one image per grapevine. 
That implies that the acquisition of images in the field is about $24$ times faster compared to making manual measurements. 
Except the provision of the captured images the framework needs no human user interaction and automatically analyze the images in order to make it available to the user. 
Thus, the analysis can be performed in parallel to the usual work within the breeding program, but also allows for a retrospective analysis. 
At the moment the program needs about $2 - 3$~minutes per image on a standard computer.

\renewcommand{\baselinestretch}{1}  
\begin{figure}[ht]
  \centering
  	 \subfigure[Riesling BBCH~75]{\includegraphics[width=0.31\textwidth]{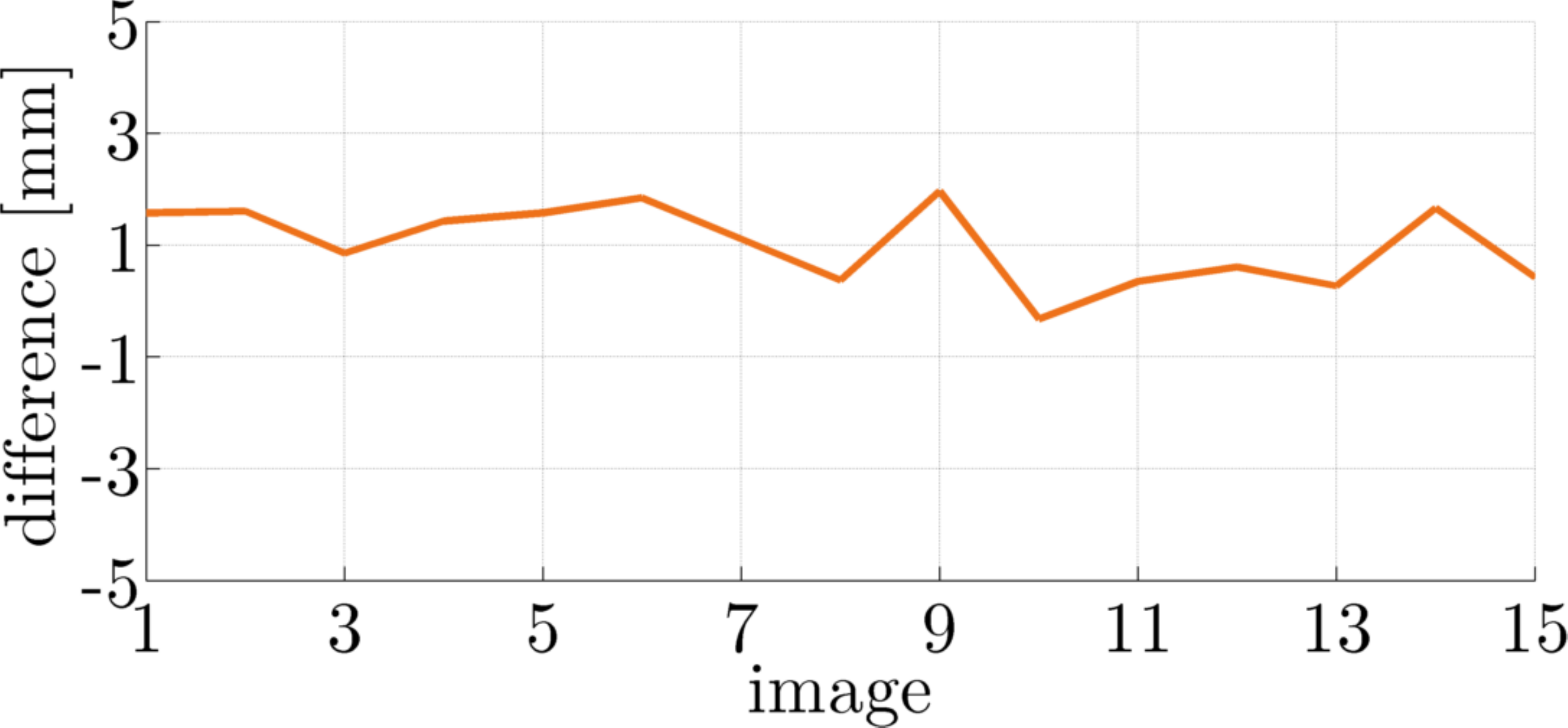}}\quad
  	 \subfigure[Riesling BBCH~81]{\includegraphics[width=0.31\textwidth]{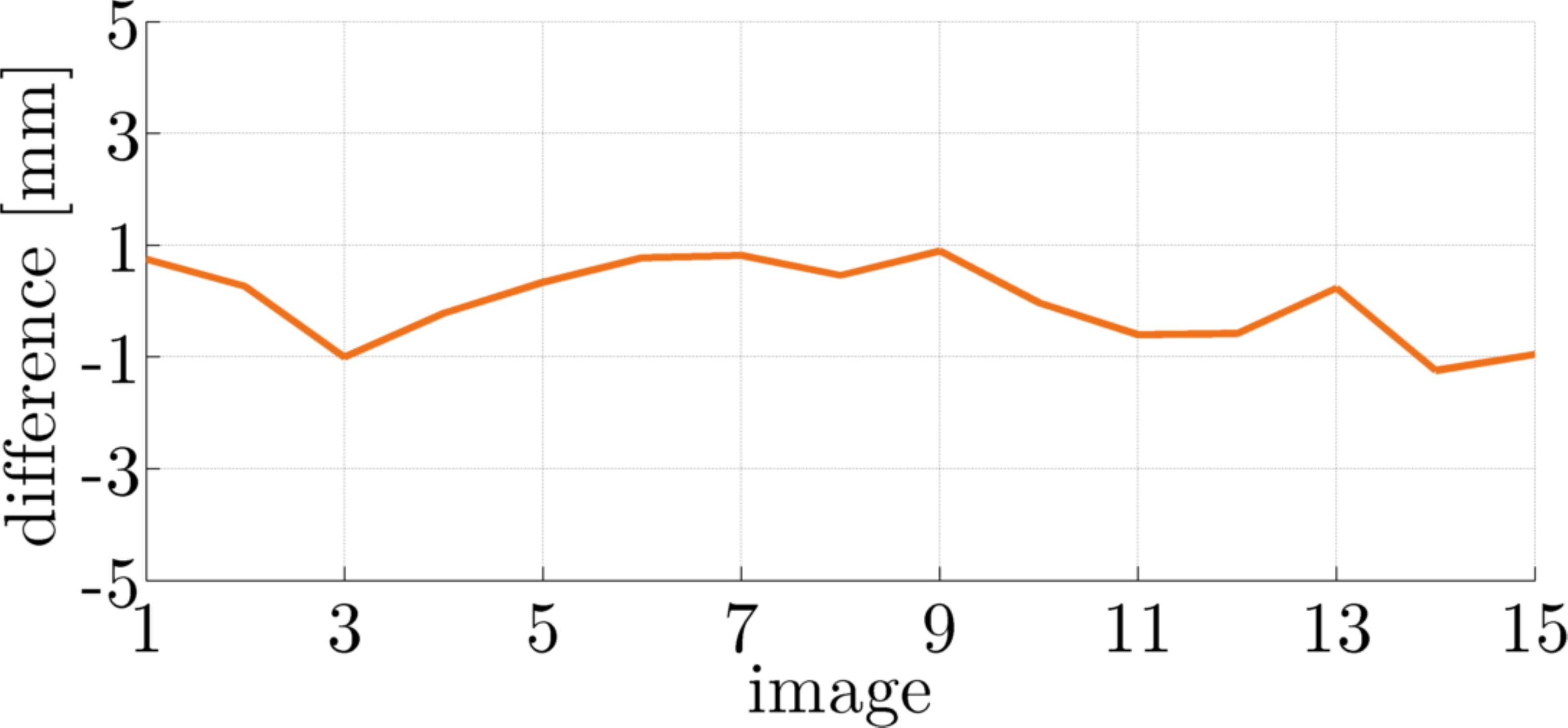}}\quad
  	 \subfigure[Riesling BBCH~89]{\includegraphics[width=0.31\textwidth]{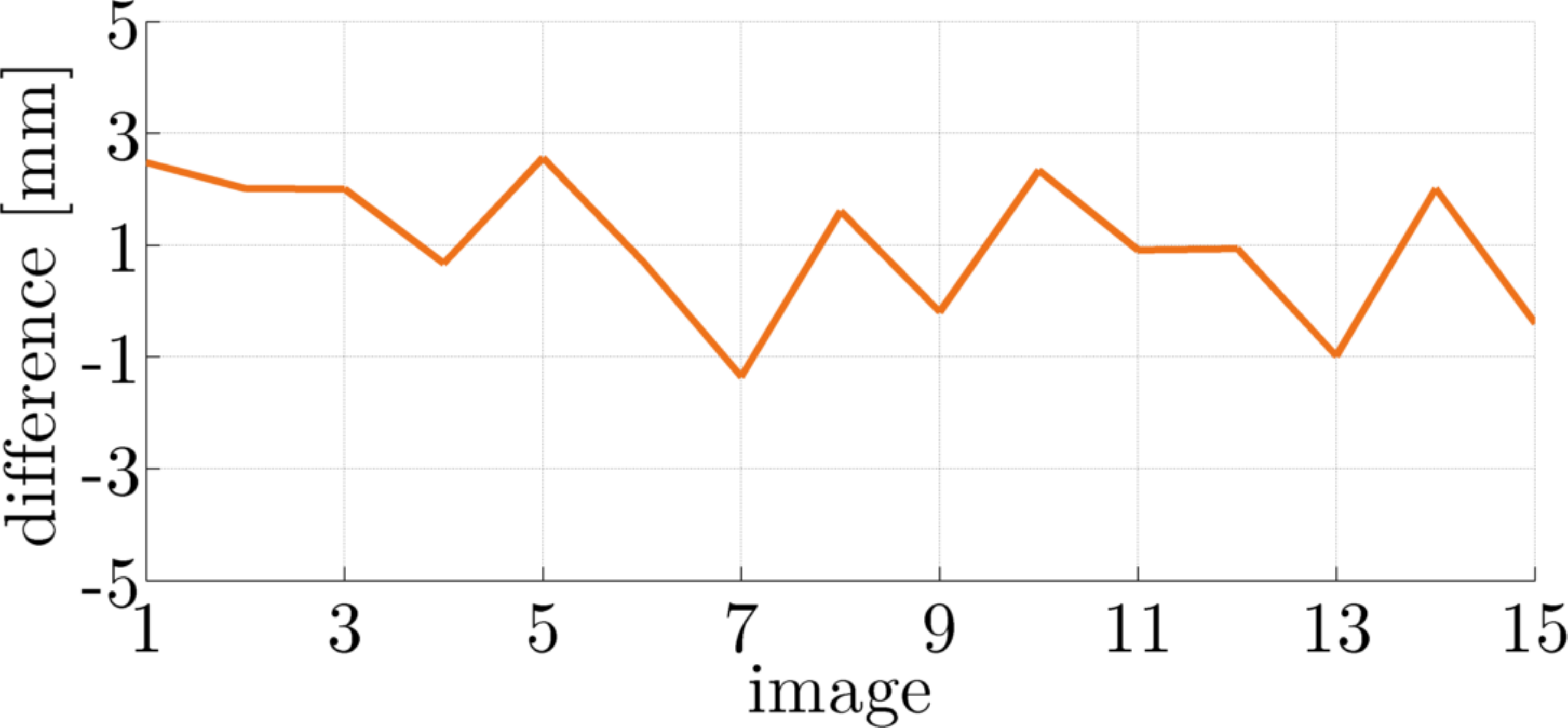}}
  	 
     \subfigure[Pinot Blanc BBCH~75]{\includegraphics[width=0.31\textwidth]{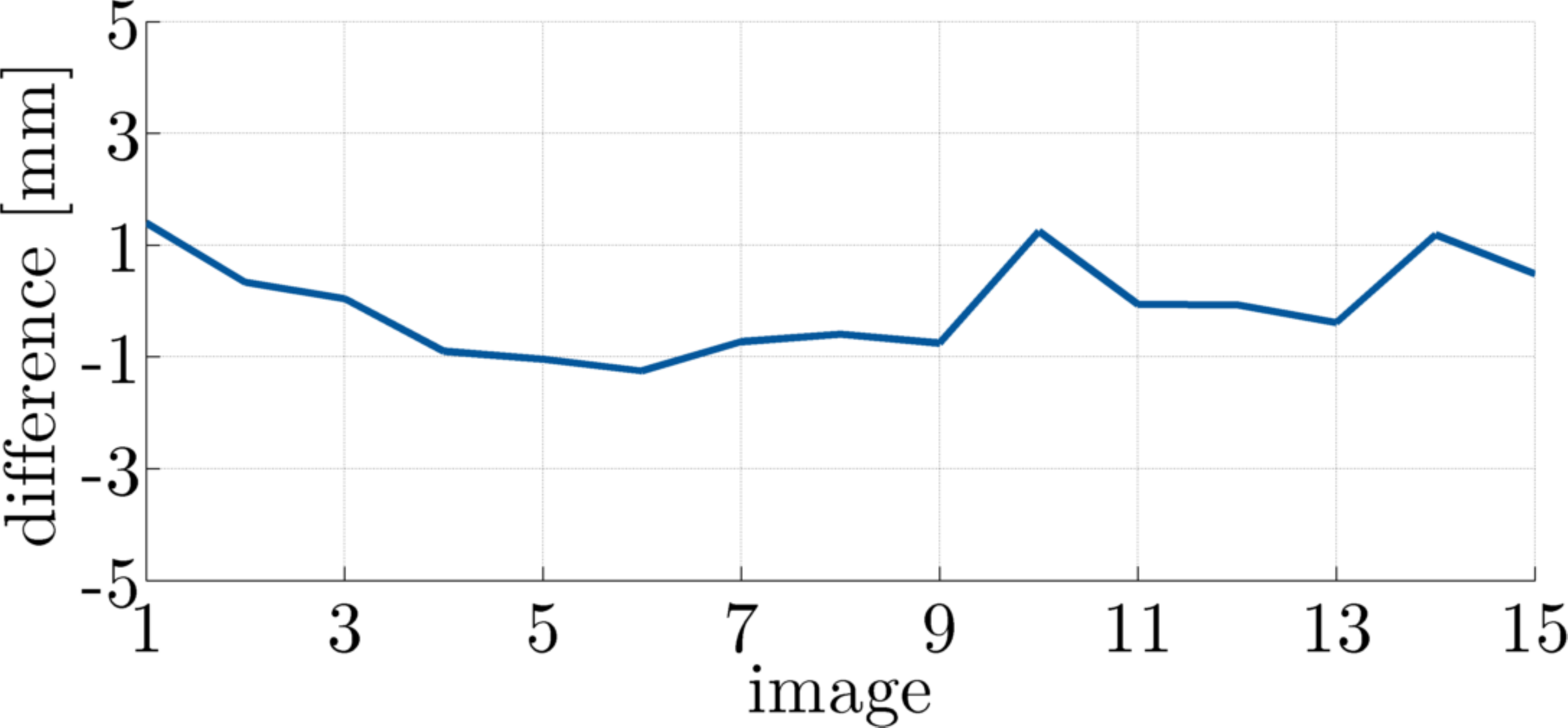}}\quad
     \subfigure[Pinot Blanc BBCH~81]{\includegraphics[width=0.31\textwidth]{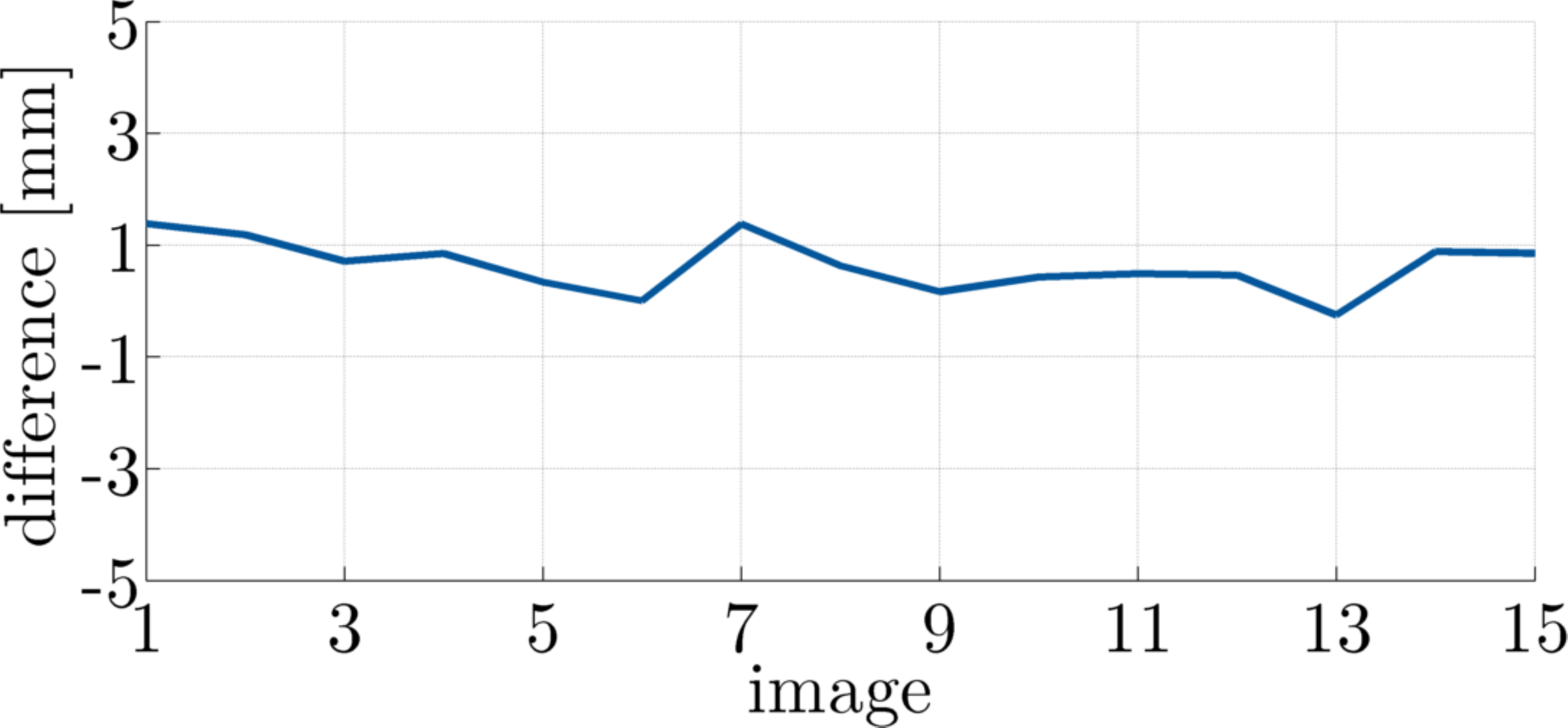}}\quad
     \subfigure[Pinot Blanc BBCH~89]{\includegraphics[width=0.31\textwidth]{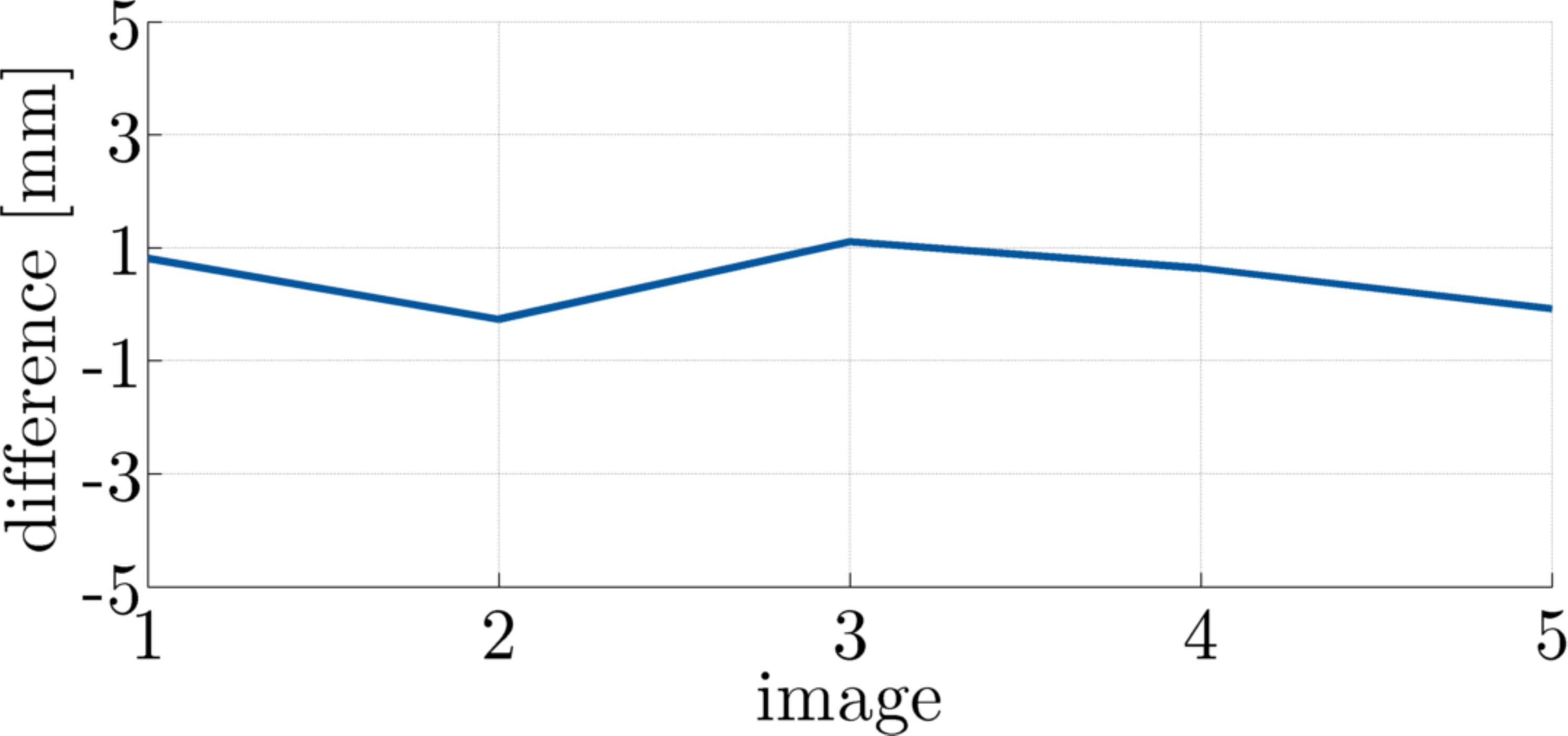}}
          
     \subfigure[Pinot Noir BBCH~75]{\includegraphics[width=0.31\textwidth]{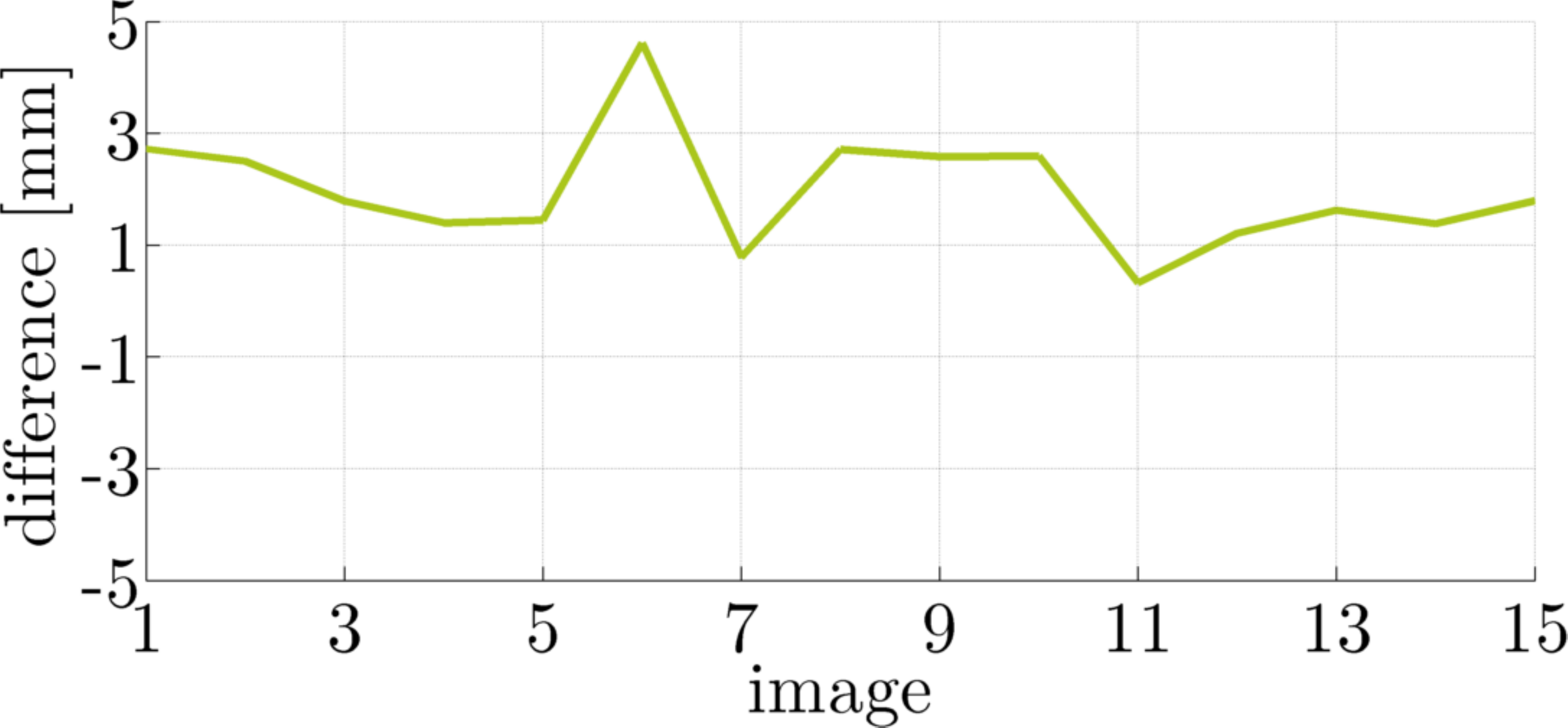}}\quad
     \subfigure[Pinot Noir BBCH~81]{\includegraphics[width=0.31\textwidth]{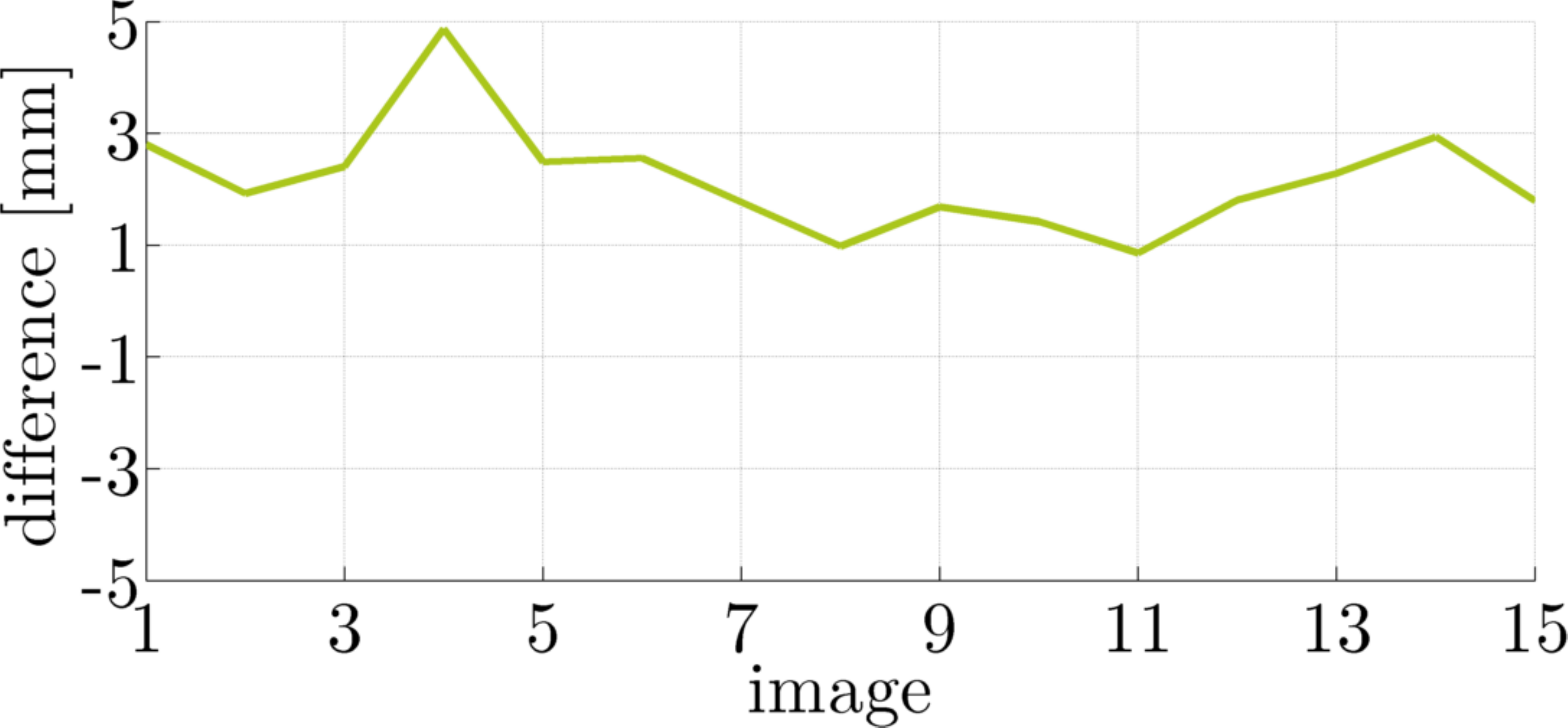}}\quad
     \subfigure[Pinot Noir BBCH~89]{\includegraphics[width=0.31\textwidth]{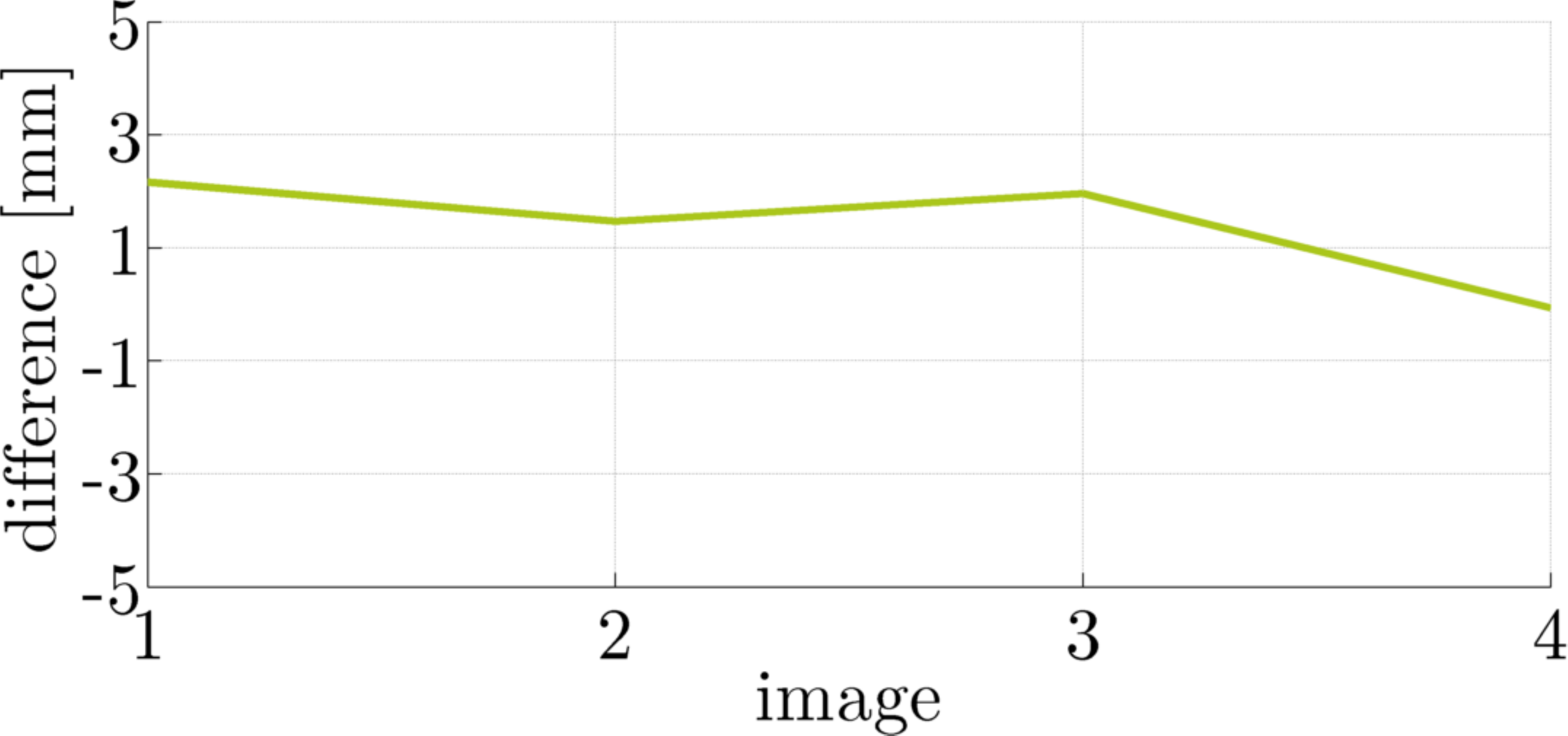}}
     
     \subfigure[Dornfelder BBCH~75]{\includegraphics[width=0.31\textwidth]{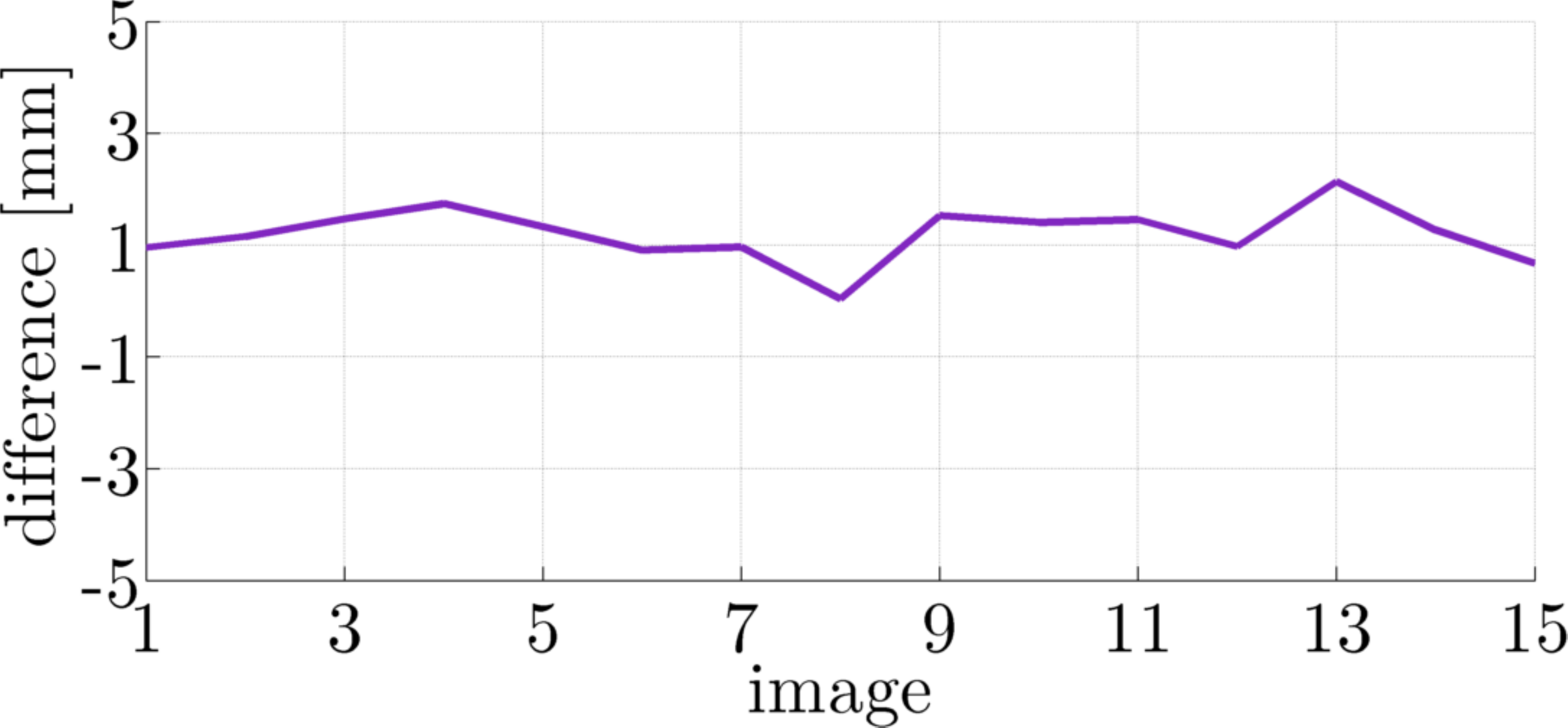}}\quad
     \subfigure[Dornfelder BBCH~81]{\includegraphics[width=0.31\textwidth]{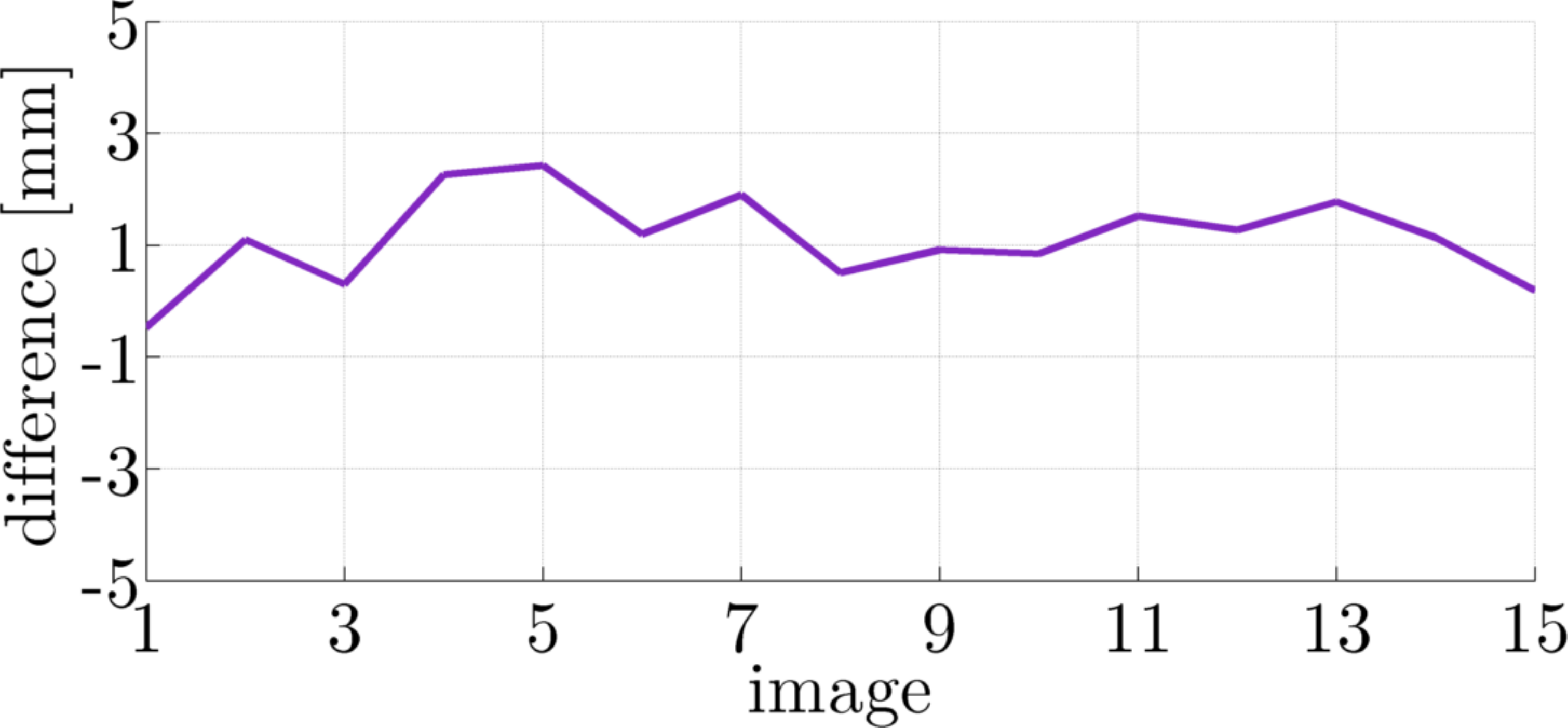}}\quad
     \subfigure[Dornfelder BBCH~89]{\includegraphics[width=0.31\textwidth]{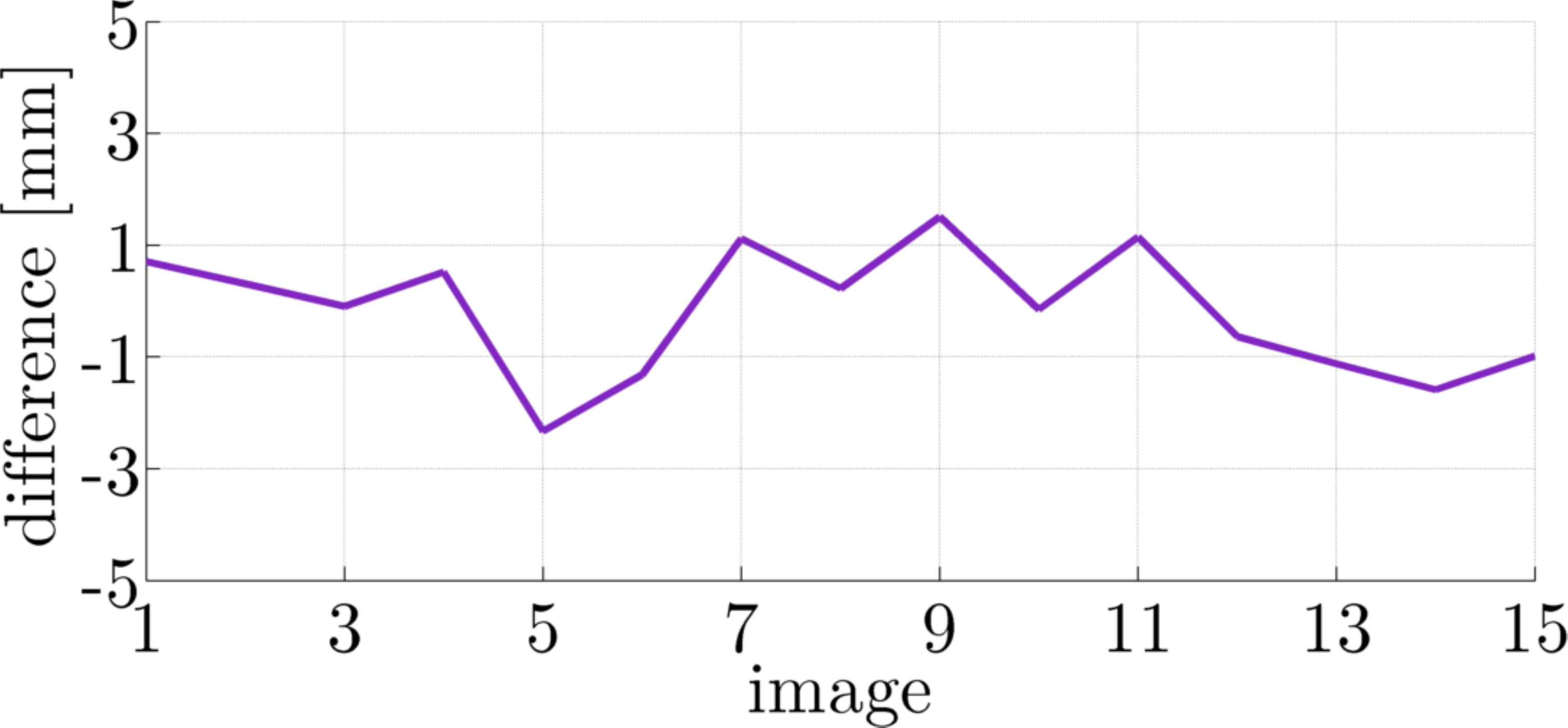}}
  \caption{
    Differences between the mean the manual measurements.}
  \label{fig:diff}
\end{figure}
\renewcommand{\baselinestretch}{2}

Figure~\ref{fig:diff} shows the differences of the estimated mean berry sizes to the manual measurements for all sorts and stages of growth.
It can be seen that the mean diameter have in most cases differences not more than $2$~mm.
The highest differences are observed when small berries are not distinct enough to be detected by the circle detector. 
The variations in the plots can be explained by the fact that each image has different conditions regarding illumination and visibility of berries.
Thus, a reliable evaluation is only guaranteed if several images of the grapevine or images of grapevines of the same sort around the same time are acquired and their results averaged. 

\renewcommand{\baselinestretch}{1}
\begin{figure}[]
  \centering
     \subfigure[Original image of Pinot Blanc (BBCH~81)]{\includegraphics[width=0.48\textwidth]{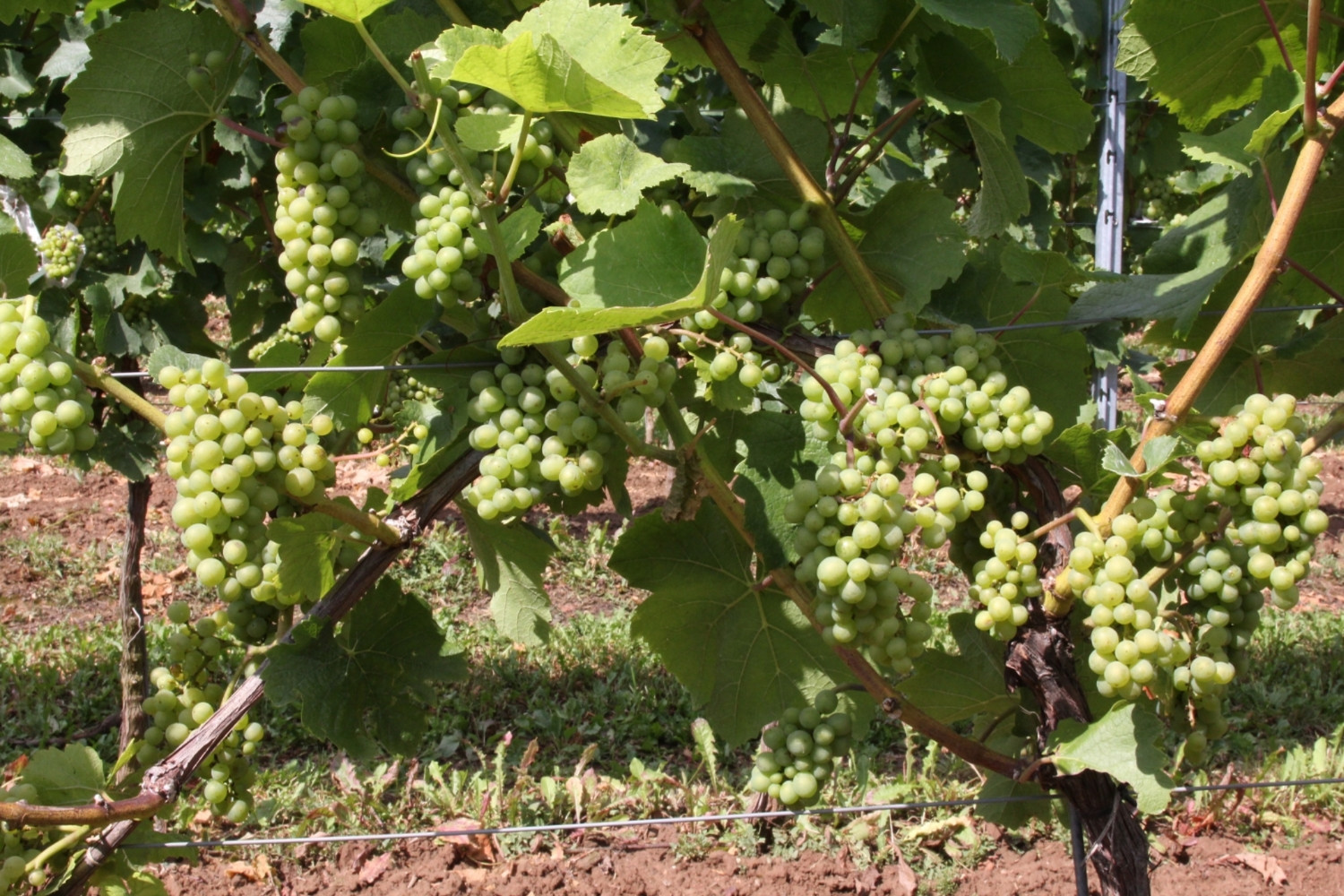}}\quad
      \subfigure[Original image of Pinot Blanc (BBCH~89)]{\includegraphics[width=0.48\textwidth]{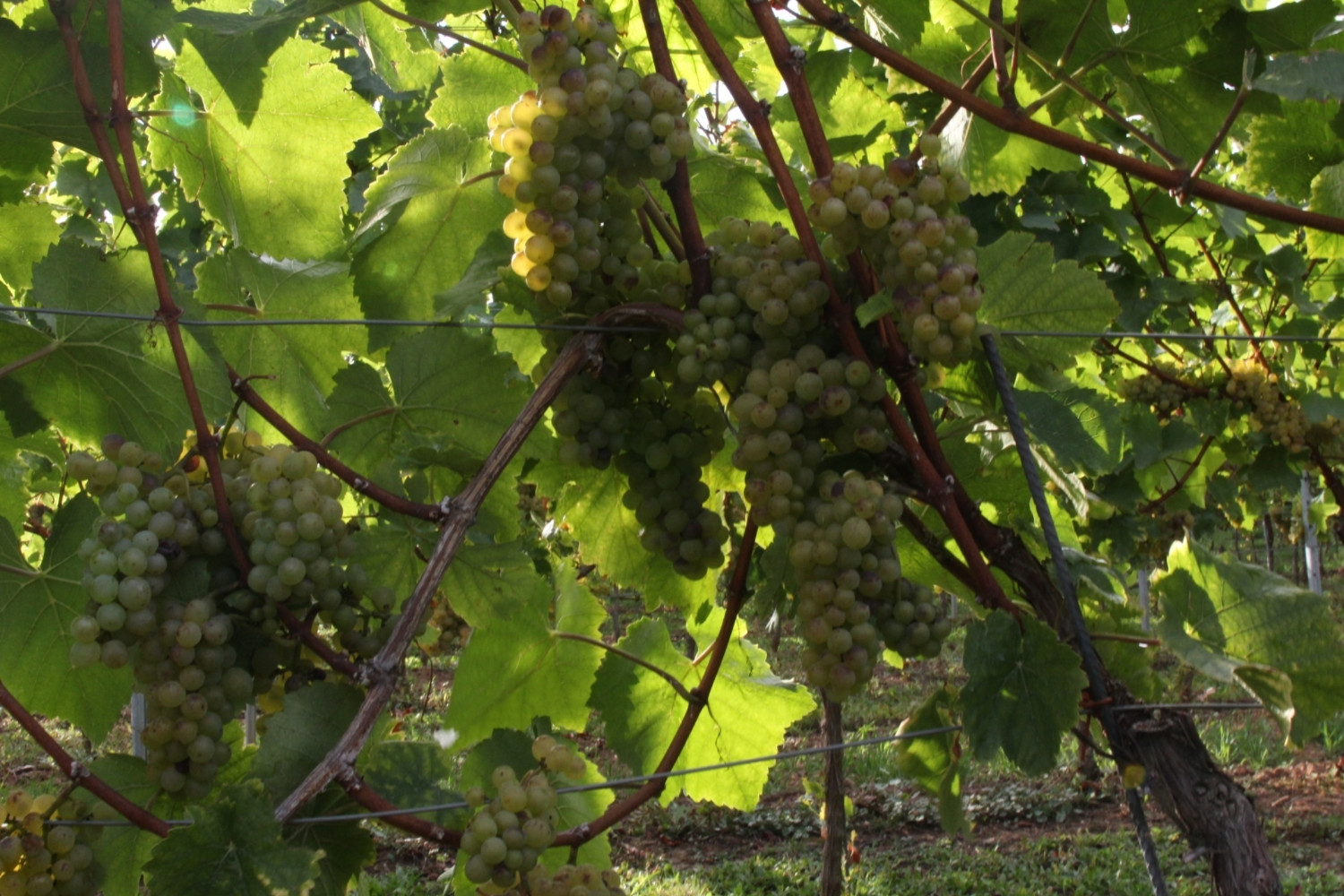}}

  	  \subfigure[Classification result of Pinot Blanc (BBCH~81) without binary term]{\includegraphics[width=0.48\textwidth]{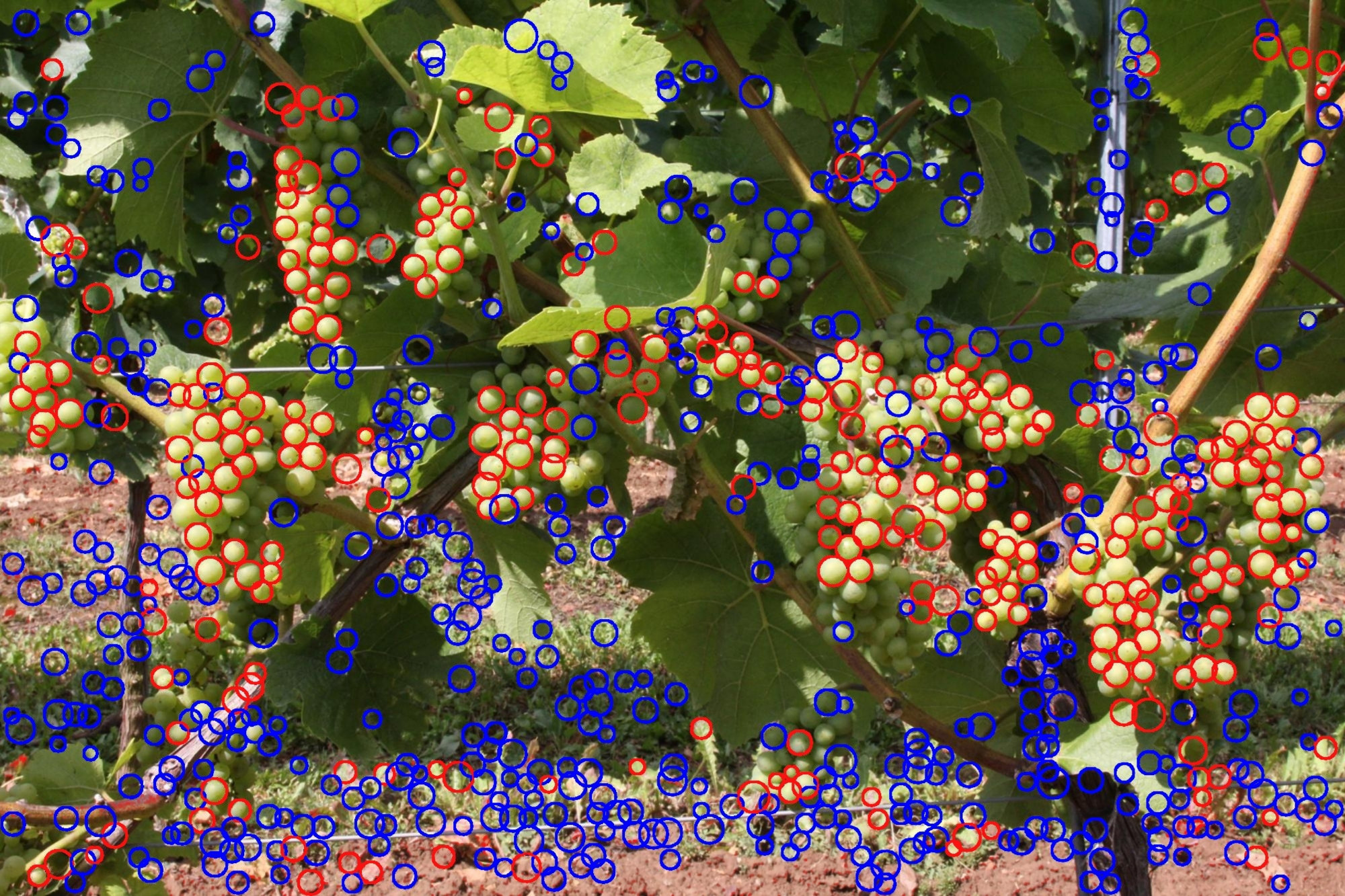}}\quad
      \subfigure[Classification result of Pinot Blanc (BBCH~89) without binary term]{\includegraphics[width=0.48\textwidth]{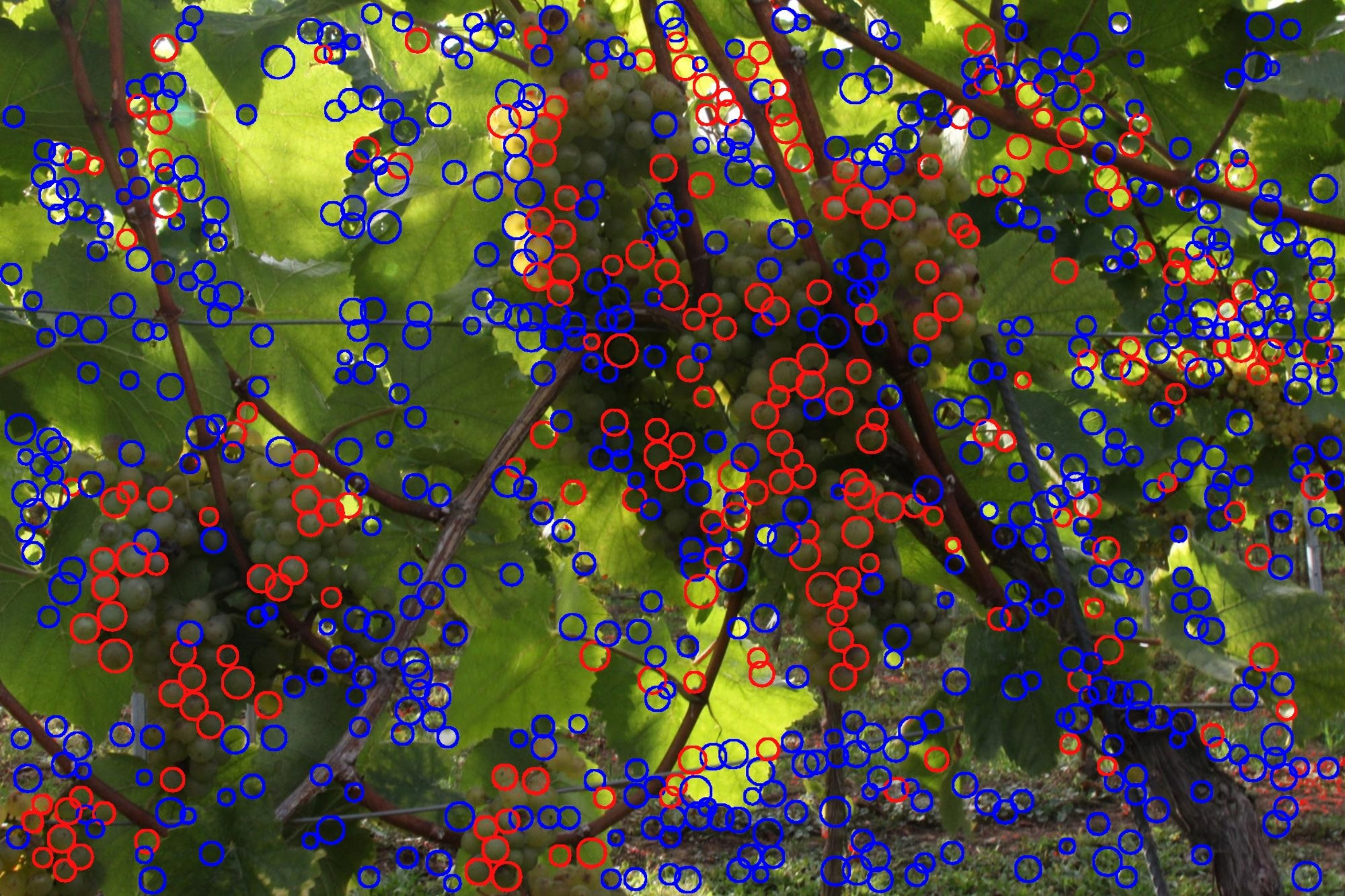}}
      
      \subfigure[Classification result of Pinot Blanc (BBCH~81) with CRF]{\includegraphics[width=0.48\textwidth]{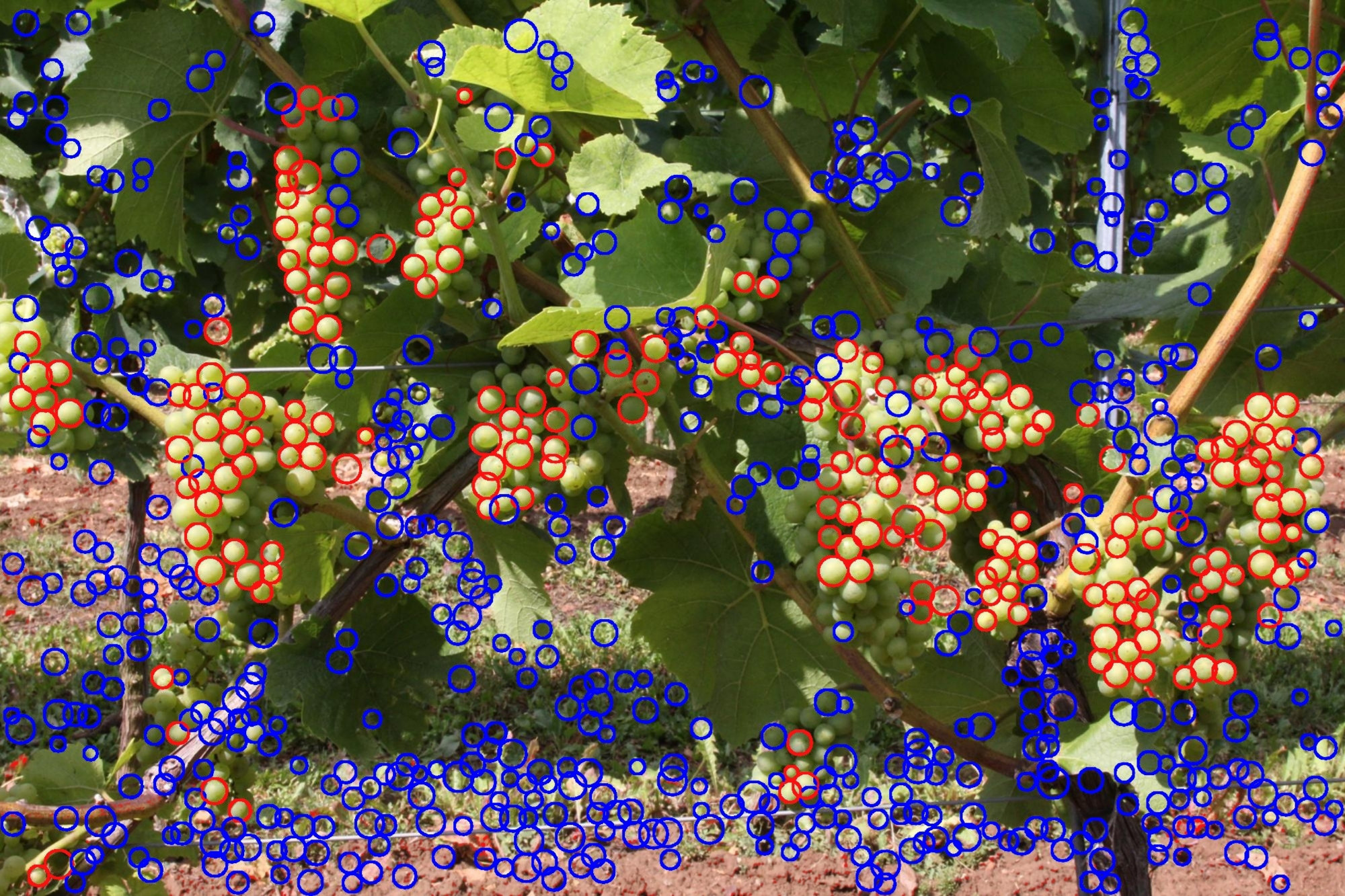}}\quad
      \subfigure[Classification result of Pinot Blanc (BBCH~89) with CRF]{\includegraphics[width=0.48\textwidth]{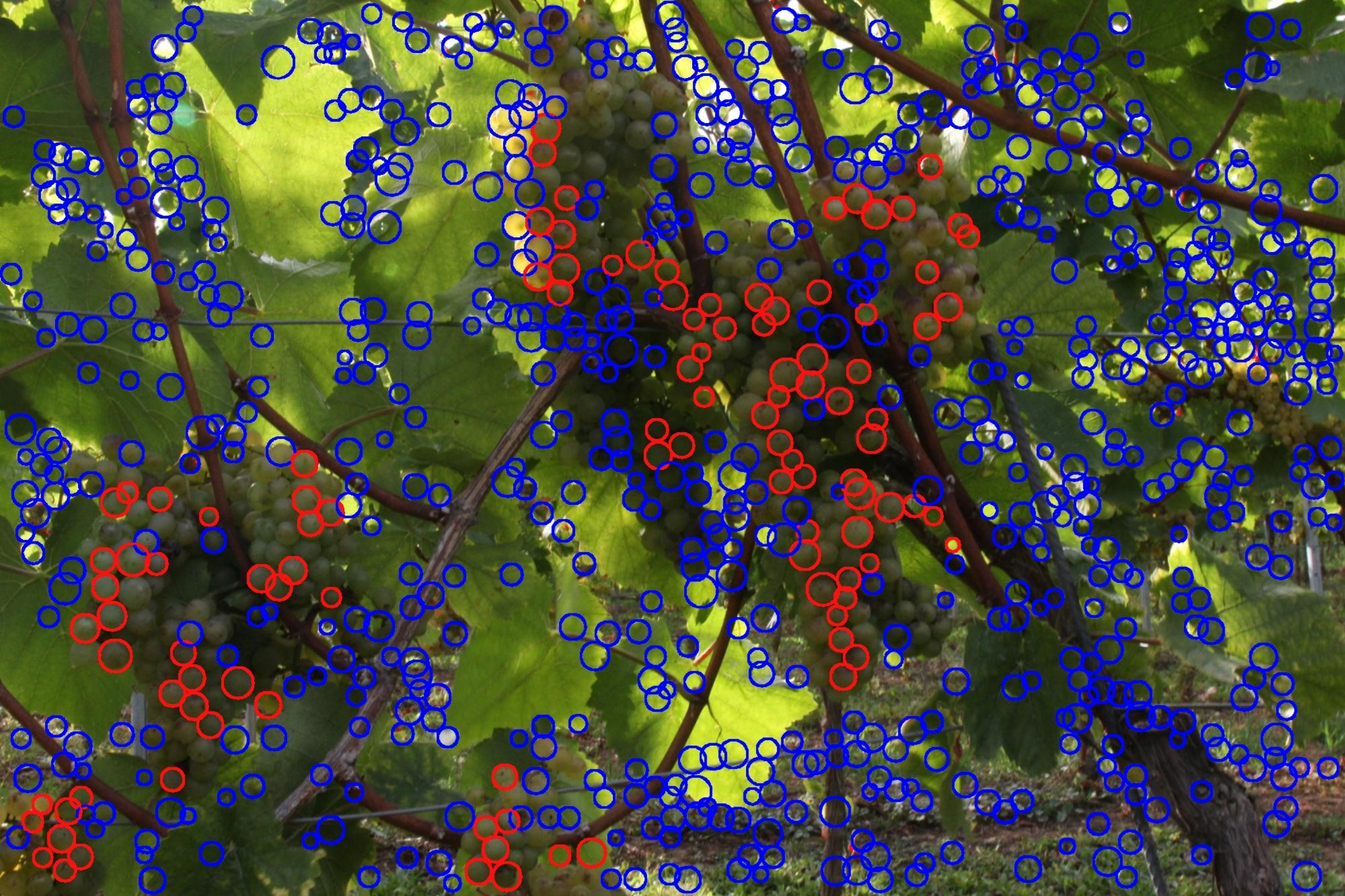}}

  \caption{
    Top row: Original images; Second and third row: Example results of candidates with position and radius (circles) and their classification: red are candidates which are classified as 'berry' and blue are candidates which are classified as 'non-berry'. Second row: Classification result without the usage of the binary term; Third row: Classification result with the usage of the CRF, \ie unary and binary term.}
  \label{fig:res_images}
\end{figure}
\renewcommand{\baselinestretch}{2}

Figure~\ref{fig:res_images} shows the classification results and the obtained berry sizes of two images of the sort 'Pinot Blanc'.
The first row shows the original image, the second row the classification result using only the unary terms without introduced prior knowledge about the arrangement of berries and the third row shows the classification result using the conditional random field.
Using the conditional random field with unary and binary terms yields slightly better results for BBCH~75 and BBCH~81, but a significantly better result for BBCH~89.
Isolated berries are eliminated and many circles which features are classified as 'non-berry' are correctly classified as 'berry' when introducing the prior knowledge about the arrangement of the berries. 
The presented results reflect the obtained classifications of the other images used in this experiments, so that the conditional random field contributes most to images showing BBCH~89.
Considering the images showing BBCH~75, the classification with and without conditional random field yield similar results.
One reason is that for BBCH~89 more reference circles as well as candidates could be found than for BBCH~75, and thus a more suitable neighborhood graph can be obtained since the assumption that neighbored candidates tend to have the class label is fulfilled best. 
Another reason is that the images of BBCH~89 were taken on sunny days, so that the appearing backlight causes many distinct structures in the background with features similar to these of berries, yielding detected circles which are incorrectly classified as 'berry'.
These incorrectly classified circles are rather spread over the whole image than clustered and thus, can be eliminated using the neighbored circles obviously classified as 'non-berry'.

\renewcommand{\baselinestretch}{1}
\begin{figure}[ht]
  \centering
      \subfigure[Berry sizes of Pinot Blanc (BBCH~81)]
{\includegraphics[width=0.48\textwidth]{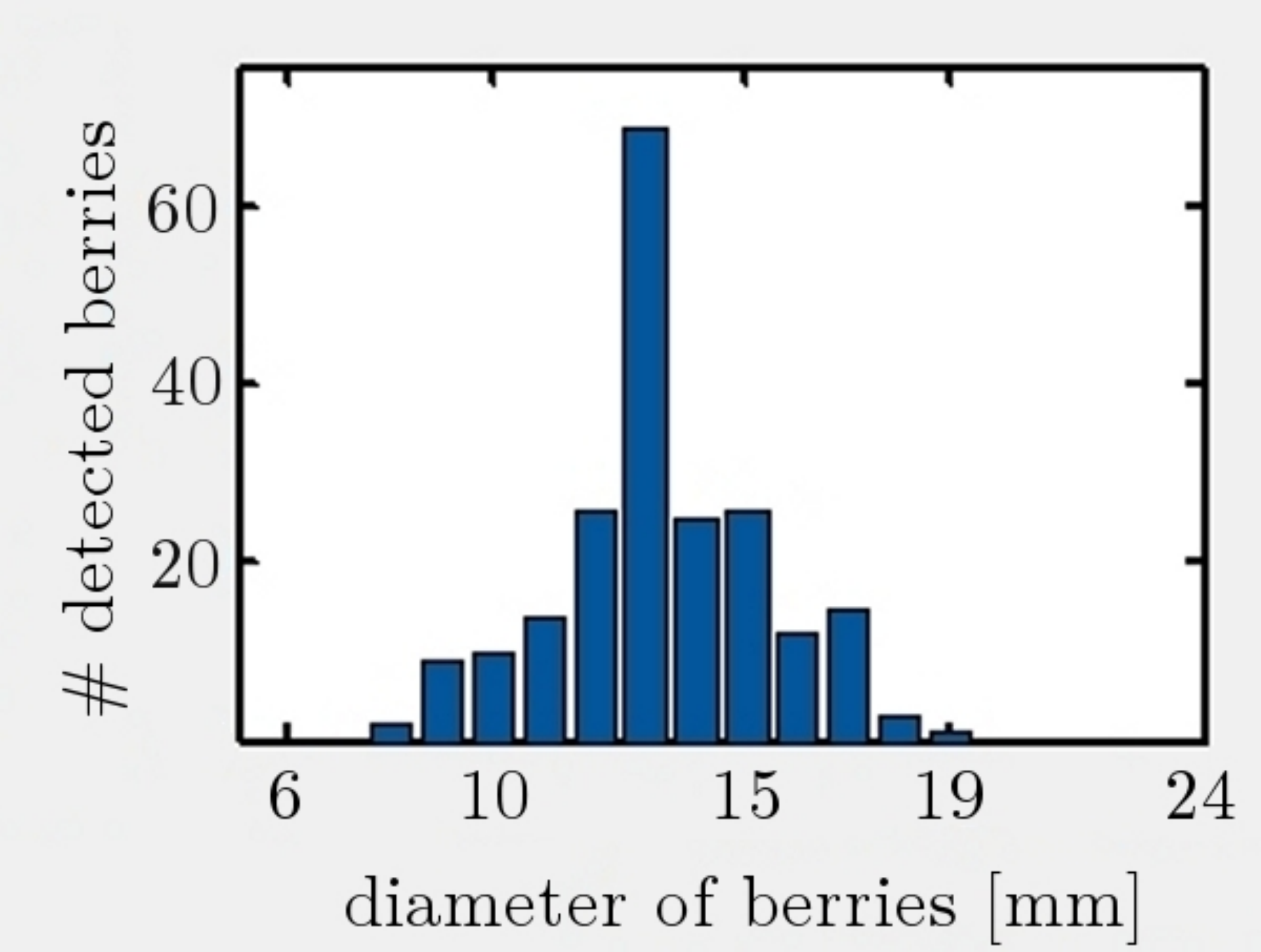}}\quad
      \subfigure[Berry sizes of  Pinot Blanc (BBCH~89)]{\includegraphics[width=0.48\textwidth]{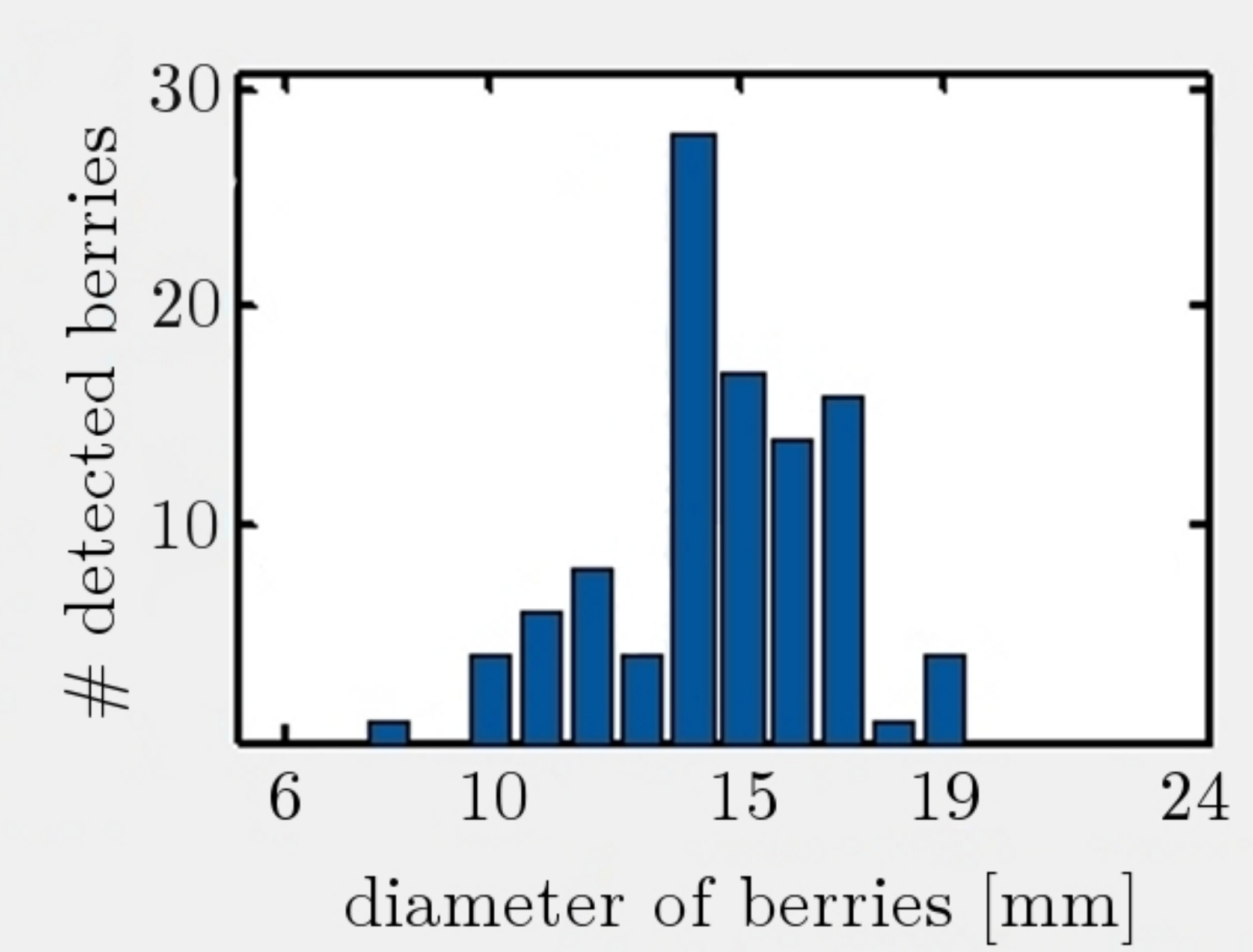}}
  \caption{Histogram of determined berry diameters. The histograms show a wide range of diameters but a concentration around a peak.}
  \label{fig:res_histo}
\end{figure}
\renewcommand{\baselinestretch}{2}

As can be seen in Figure \ref{fig:res_histo} all histograms show a wide range of found diameters, but a concentration around the highest peak.
Because the development of the berries vary within each cluster and thus, also in the vineyard, the histogram allows for a more meaningful interpretation than only the mean of the detected berries.

A basic assumption of the approach states that the silhouette of the berries is characterized by strong image gradients.
Violations of this assumptions cause the circle detection module to fail 
to proper detect the berry's boundaries or to define a suitable reference set, which clearly represents the major limitation of the approach.
The problem can be overcome by using additional lights when taking the images.

In some cases it cannot be guarenteed that enough berries are visible in the image, \eg if the image acquisition is fully automated.
In this case the detection step for reference circles fails or gives only poor results.
This step can be replaced by the usage of manually selected berries acquired from other images.
However, the selected training data needs to be appropriate in order to be representative for the candidates.
In order to proof the applicability of this approach, the experiments presented here were also conducted with 150 manually selected green berries of BBCH~81, which were used as reference berries.
Using only gist features yields the best results with mean absolute differences ranging from $0.3$~mm for 'Pinot Blanc' and 'Dornfelder' to $1.6$~mm for 'Pinot Noir' in BBCH~75, from $0.6$~mm for 'Pinot Blanc' and 'Pinot Noir' to $1.0$~mm for 'Dornfelder' in BBCH~81 and from $0.6$~mm for 'Pinot Noir' to $2.9$~mm for 'Dornfelder' in BBCH~89.
The results deteriorates for larger berry sizes, because the chosen reference berries were collected from an earlier stage. 
To overcome this problem the reference berries should be roughly chosen according to the current berry size.
Nevertheless, the approach seems promising due to the fact that even dark berries could be detected using green reference berries.

\begin{figure}[ht]
  \centering
  	 \subfigure[One image captured with camerasystem]{\includegraphics[height=0.24\textheight]{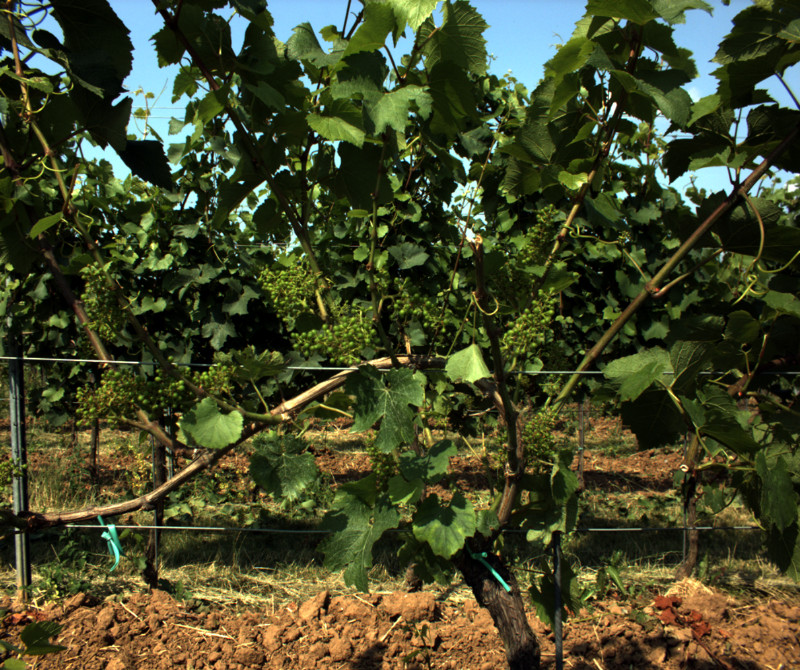}}\quad
  	 \subfigure[Computed depthmap]{\includegraphics[height=0.24\textheight]{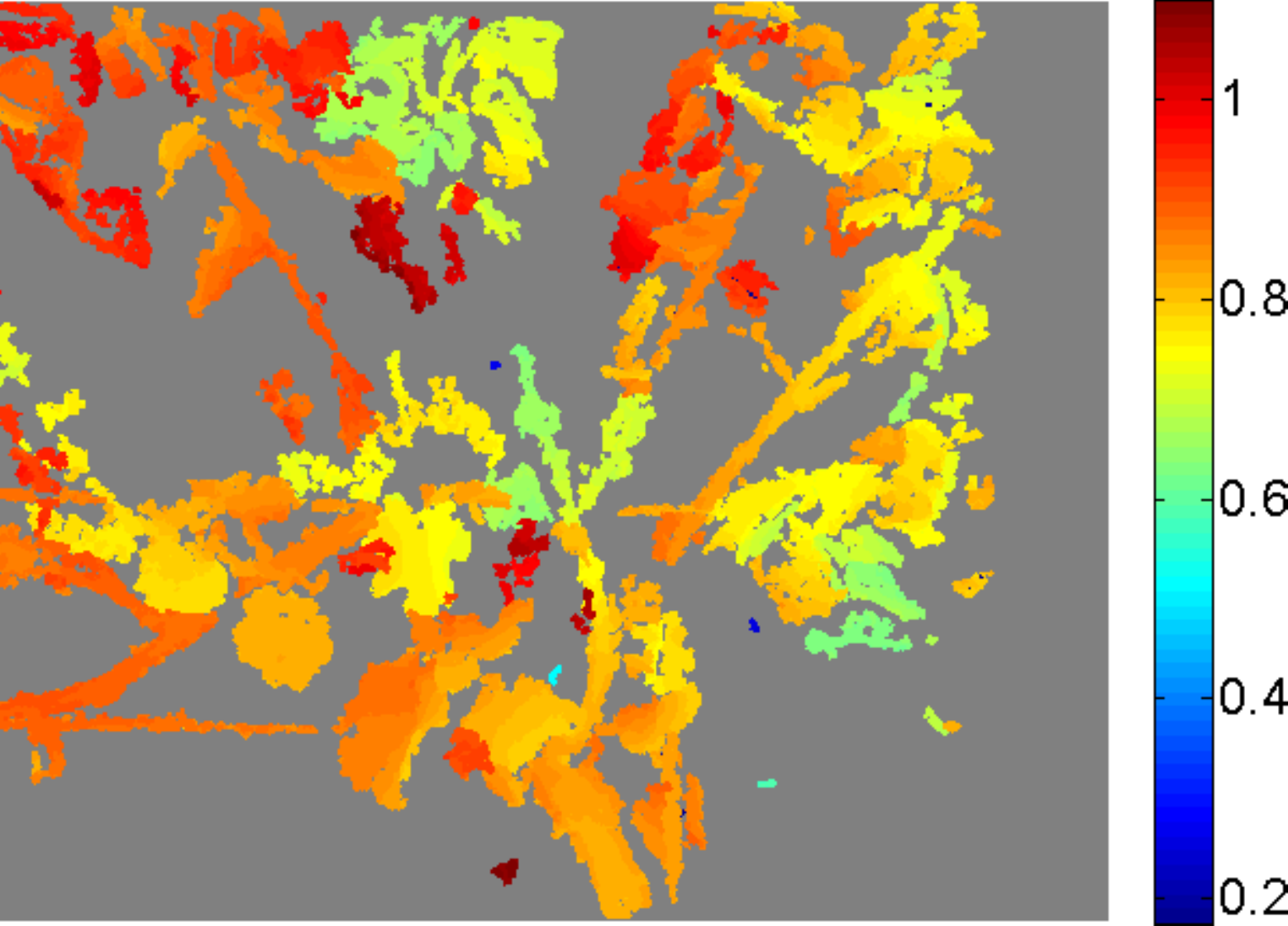}}
  \caption{Left: One image from a camera system with known interior and relative orientation. Right: The depth information, given in m, is color-coded, whereas red pixels indicate large distances, blue pixels small distances and gray pixels indicate void distances, which can be assumed to be background.}
  \label{fig:depth}
\end{figure}

A shortcoming of the proposed framework is the conversion from pixel to mm.
It can be defined more accurately when using a camerasystem with known interior and relative orientation.
In contrast to the current approach, where only one scale for the whole image is used, a depth map can be derived in order to define a pixelwise scale.
First experiments regarding the computation of depth maps are showing promising result since the approach assumes a 3D architecture of the grapevine rather than a single plane (see Figure \ref{fig:depth}).
The depth maps were computed using patch-based multi-view stereo software proposed by \cite{Furukawa2010} and the orientation was obtained using the approach of \cite{Abraham1997}.
The depth maps would also enable the removal of far-off background in order to restrict the sets of circles to those circles lying in a distance of about $1$m.

\renewcommand{\baselinestretch}{1}  
\renewcommand{\baselinestretch}{2}

\section{Conclusion}
\label{sec:conclusion}
The paper proposed a high-throughput image analysis framework, which non-invasively detect grapevine berries and determine their size in mm from RGB images which were taken in vineyards.
The framework automatically detects berries in images by first finding circular structures and classify them into the classes 'berry' and 'non-berry' using the concept of one-class classification.
The classification is done by utilizing an active learning framework and a conditional random field.
The experiments could show that the framework is able to detect a representative amount of berries in order to extract a reliable quantity of phenotypic data while keeping the error rate of falsely detected berries as low as possible in order to ensure a high quality of the extracted data.
The obtained results showed a mean difference of about $1$~mm to manual reference measurements and a correlation between the mean berry size and the manual reference measurements by $0.88$.

\bibliographystyle{elsarticle-harv}

\end{document}